\PassOptionsToPackage{table}{xcolor}

\documentclass[sigconf]{acmart}

\AtBeginDocument{%
  }

\usepackage{subcaption}
\usepackage{multirow}
\usepackage{booktabs}
\usepackage{xcolor}

\definecolor{risklow}{RGB}{199,233,192}
\definecolor{riskmed}{RGB}{255,237,160}
\definecolor{riskhigh}{RGB}{252,146,114}
\definecolor{riskcrit}{RGB}{222,45,38}
\newcommand{\riskcell}[2]{%
  \ifdim #1 pt > 0.89 pt
    \cellcolor{riskcrit}\textcolor{white}{\textbf{#2}}%
  \else\ifdim #1 pt > 0.59 pt
    \cellcolor{riskhigh}#2%
  \else\ifdim #1 pt > 0.29 pt
    \cellcolor{riskmed}#2%
  \else
    \cellcolor{risklow}#2%
  \fi\fi\fi
}

\copyrightyear{2026}
\acmYear{2026}
\setcopyright{cc}
\setcctype{by}
\acmConference[BCB '26]{17th ACM International Conference on Bioinformatics, Computational Biology and Health Informatics}{June 30-July 03, 2026}{Rende (CS), Italy}
\acmBooktitle{17th ACM International Conference on Bioinformatics, Computational Biology and Health Informatics (BCB '26), June 30-July 03, 2026, Rende (CS), Italy}
\acmDOI{10.1145/3807503.3819470}
\acmISBN{979-8-4007-2653-8/2026/06}

\begin{document}
\setlength{\textfloatsep}{6pt plus 2pt minus 2pt}

\title{Quantifying Memorization and Privacy Risks in Genomic Language Models}

\author{Alexander Nemecek}
\affiliation{%
  \institution{Case Western Reserve University}
  \city{Cleveland}
  \state{Ohio}
  \country{USA}
}

\author{Wenbiao Li}
\affiliation{%
  \institution{Case Western Reserve University}
  \city{Cleveland}
  \state{Ohio}
  \country{USA}
}

\author{Xiaoqian Jiang}
\affiliation{%
  \institution{The University of Texas Health Science Center at Houston}
  \city{Houston}
  \state{Texas}
  \country{USA}
}

\author{Jaideep Vaidya}
\affiliation{%
 \institution{Rutgers University}
 \city{Newark}
  \state{New Jersey}
  \country{USA}
}

\author{Erman Ayday}\authornote{Correspondence: \href{mailto:exa208@case.edu}{exa208@case.edu}.}
\affiliation{%
  \institution{Case Western Reserve University}
  \city{Cleveland}
  \state{Ohio}
  \country{USA}
}

\renewcommand{\shortauthors}{Nemecek et al.}

\begin{abstract}
Genomic language models (GLMs) have emerged as powerful tools for learning representations of DNA sequences, enabling advances in variant prediction, regulatory element identification, and cross-task transfer learning. However, as these models are increasingly trained or fine-tuned on sensitive genomic cohorts, they risk memorizing specific sequences from their training data, raising serious concerns around privacy, data leakage, and regulatory compliance. Despite growing awareness of memorization risks in general-purpose language models, little systematic evaluation exists for these risks in the genomic domain, where data exhibit unique properties such as a fixed nucleotide alphabet, strong biological structure, and individual identifiability. We present a comprehensive, multi-vector privacy evaluation framework designed to quantify memorization risks in GLMs. Our approach integrates three complementary risk assessment methodologies: perplexity-based detection, canary sequence extraction, and membership inference. These are combined into a unified evaluation pipeline that produces a worst-case memorization risk score. To enable controlled evaluation, we plant canary sequences at varying repetition rates into both synthetic and real genomic datasets, allowing precise quantification of how repetition and training dynamics influence memorization. We evaluate our framework across multiple GLM architectures, examining the relationship between sequence repetition, model capacity, and memorization risk. Our results establish that GLMs exhibit measurable memorization and that the degree of memorization varies across architectures and training regimes. These findings reveal that no single attack vector captures the full scope of memorization risk, underscoring the need for multi-vector privacy auditing as a standard practice for genomic AI systems. All code is available at https://github.com/ANCP2021/GLM-Memorization.
\end{abstract}

\ccsdesc[500]{Security and privacy~Privacy protections}
\ccsdesc[300]{Computing methodologies~Natural language processing}
\ccsdesc[300]{Applied computing~Computational genomics}

\keywords{Genomic Language Models, Memorization, Membership Inference Attacks, Privacy}


\maketitle

\section{Introduction}
\label{sec:introduction}

The rapid expansion of large-scale genomics has produced unprecedented volumes of sequencing data, creating both opportunities and challenges for computational biology. In parallel, the success of self-supervised pretraining in natural language processing has demonstrated that large language models (LLMs) can learn rich representations from unlabeled text through objectives such as masked language modeling and autoregressive next-token prediction. Genomic language models (GLMs) adapt this paradigm to DNA and RNA sequences, treating nucleotide strings as biological ``texts'' governed by an underlying grammar of regulatory logic, evolutionary constraint, and functional encoding. The field has advanced rapidly through architectures spanning masked encoders, long-range convolutional models, and state-space architectures, with parameter counts growing from 110 million to 40 billion~\cite{dallatorre2023nucleotide, zhou2024dnabert2, nguyen2023hyenadna, nguyen2024evo}. As a result, GLMs are being incorporated into research pipelines and increasingly explored for clinical applications, including variant effect prediction, regulatory element identification, and cross-task transfer learning. However, as these models move toward deployment in high-stakes biomedical settings, particularly when fine-tuned on smaller, domain-specific genomic cohorts, concerns around privacy, data leakage, and regulatory compliance have emerged as barriers to responsible adoption.

Three properties of genomic data make memorization uniquely consequential: sequences are immutable and cannot be reissued once compromised~\cite{erlich2018identity, gymrek2013identifying}, identifiable from as few as several hundred variants~\cite{homer2008resolving, erlich2014routes}, and heritable, exposing biological relatives who never consented to data collection. Memorization in LLMs follows well-characterized scaling laws tied to model capacity, data duplication, and context length~\cite{carlini2021extracting, carlini2019secret, carlini2023quantifying}, yet no systematic framework evaluates these risks in GLMs (Section~\ref{sec:background}).

\begin{figure}[t]
\centering
\includegraphics[width=\columnwidth]{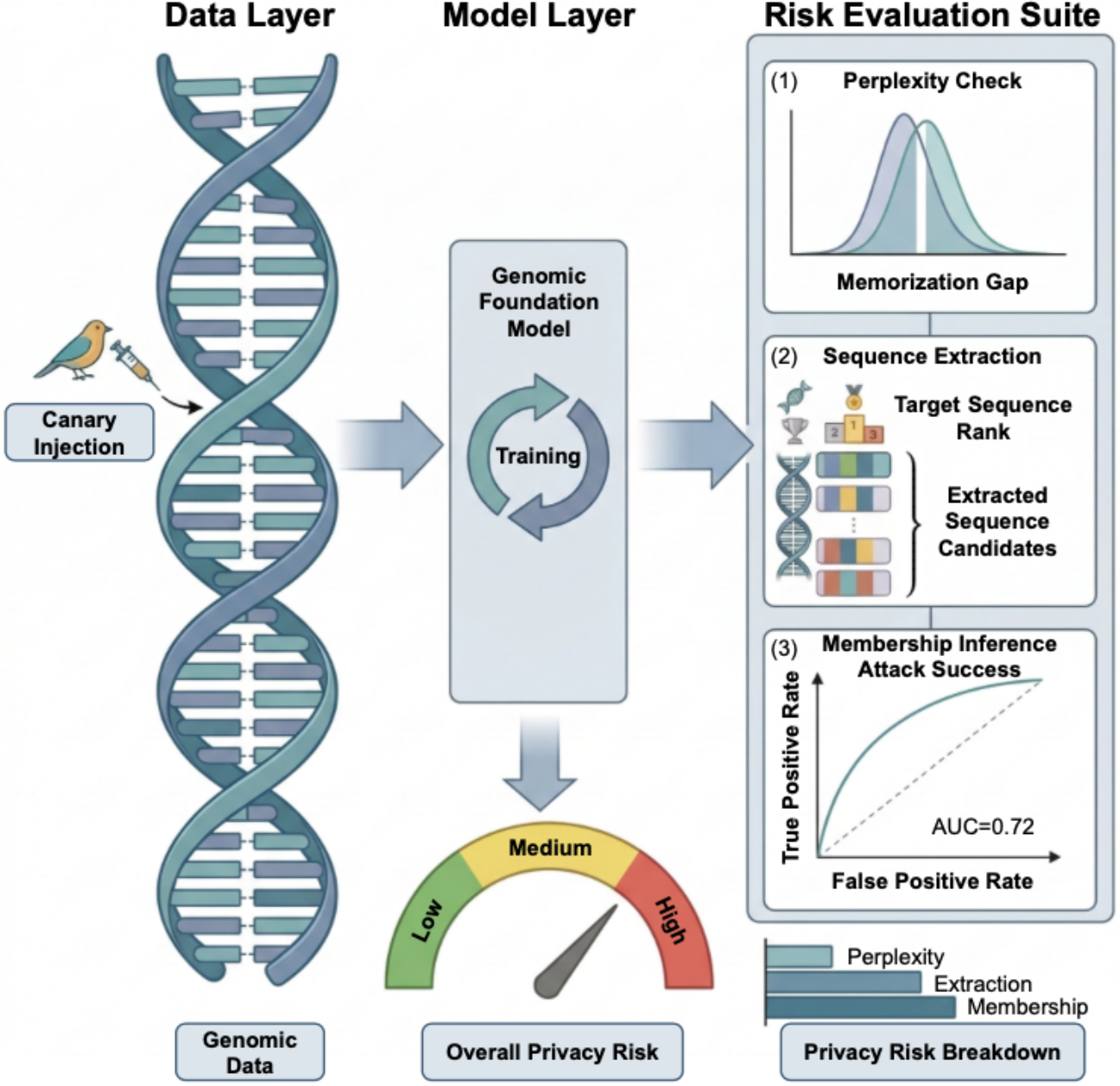}
\caption{Overview of the memorization risk quantification framework.
Genomic training data is augmented with canary sequences at controlled
repetition rates (left), used to fine-tune a genomic language model
(center), and evaluated through three complementary vectors:
perplexity-based detection, canary sequence extraction, and membership
inference (right). The outputs are combined into a maximum vulnerability
score (bottom).}
\label{fig:overview}
\end{figure}

In this paper, we present a systematic experimental study of memorization risks in GLMs. We develop a multi-vector risk quantification framework that unifies three complementary evaluation vectors: (i) perplexity-based detection, (ii) canary sequence extraction, and (iii) membership inference, into a maximum vulnerability score that characterizes worst-case memorization risk across all attack vectors for any given model-dataset configuration. Figure~\ref{fig:overview} illustrates the end-to-end pipeline. To enable controlled evaluation, we plant canary sequences at varying repetition rates into the training corpus, allowing precise quantification of how data duplication and training dynamics influence memorization.

We evaluate the framework across four GLM architectures that collectively span the principal design paradigms in the fields of masked language modeling~\cite{zhou2024dnabert2}, long-range convolutional modeling~\cite{nguyen2023hyenadna}, state-space modeling~\cite{nguyen2024evo}, and a custom lightweight transformer baseline under both full fine-tuning and parameter-efficient adaptation. Experiments are conducted across four genomic datasets of increasing biological complexity, from synthetic zero-order sequences through prokaryotic and eukaryotic reference genomes to curated multi-species benchmark data. Our results establish that all evaluated architectures exhibit measurable memorization under standard fine-tuning, that memorization scaling laws identified in natural language models transfer to the genomic domain, and that different architectures reveal memorization through different vectors. These findings demonstrate that single-metric evaluations can systematically underestimate privacy exposure and establish the proposed framework as a foundation for multi-vector privacy auditing of genomic AI.

\section{Background and Related Work}\label{sec:background}
We review memorization in language models, membership inference attacks, and privacy risks specific to genomic data.
 
\subsection{Memorization and Data Extraction in Language Models}
The empirical foundations of memorization research in language models rest on three key results by Carlini et al.~\cite{carlini2021extracting, carlini2019secret, carlini2023quantifying}: that training data is extractable via black-box access, that such memorization arises early and persists despite regularization, and that it follows log-linear scaling with model size, duplication, and prompt length. While these results were established in the context of natural language, their implications for genomic models remain largely unexplored. We focus here on subsequent work that motivates domain-specific investigation.

Training dynamics beyond model size play a large role in determining what a model memorizes. The number of training epochs, learning rate schedule, and checkpoint selection all modulate memorization behavior, with longer training generally increasing the volume of recoverable training data~\cite{carlini2019secret}. Among practical mitigations, training data deduplication has proven particularly effective. Lee et al.~\cite{lee2022dedup} showed that deduplication reduces memorized text emission by an order of magnitude and eliminates verbatim copying present in more than one percent of unprompted model outputs. Kandpal et al.~\cite{kandpal2022dedup_privacy} complemented this finding by demonstrating that deduplication directly reduces the success rate of extraction attacks, establishing a concrete link between data curation and privacy risk. These deduplication results motivate our canary insertion protocol (Section~\ref{canary-design}), which systematically varies repetition frequency to measure how duplication drives memorization in the genomic setting.

On the defense side, Carlini et al.~\cite{carlini2019secret} found that among the techniques evaluated, including early stopping, dropout, and L2 regularization, only differential privacy provided a formal guarantee against memorization, though at the cost of reduced model utility. Li et al.~\cite{li2022dp_learners} subsequently showed that large pretrained language models can serve as effective differentially private learners. These findings inform our inclusion of a LoRA-adapted model (Evo) as an experimental condition, enabling observation of whether parameter-efficient adaptation alters memorization dynamics relative to full fine-tuning.

Despite the maturity of this literature, the studies above focus exclusively on natural language text. GLM training corpora are orders of magnitude smaller and more structured, the vocabulary is constrained to four nucleotides, and individual sequences may carry identifiable information under conditions that may alter memorization dynamics. We hypothesize that duplication-driven memorization scaling laws transfer to the genomic domain, and that GLMs fine-tuned on small cohorts exhibit measurable memorization across multiple vectors. Our framework tests these hypotheses.

\subsection{Membership Inference Attacks}
\label{sec:mia-rw}
Membership inference attacks (MIAs) determine whether a specific data record was included in a model's training set. Shokri et al.~\cite{shokri2017membership} introduced the shadow training paradigm, in which auxiliary models are trained to distinguish members from non-members. Yeom et al.~\cite{yeom2018privacy} formalized the connection between overfitting and membership inference, proving that overfitting is sufficient but not necessary for successful MIA; models can leak membership information even in the absence of measurable overfitting.

Carlini et al.~\cite{carlini2022lira} advanced the state of the art with the Likelihood Ratio Attack (LiRA), which casts membership inference as a hypothesis test comparing target model behavior against a population of reference models. LiRA achieves an order-of-magnitude improvement in true-positive rate at low false-positive rates. This evaluation criterion is particularly relevant to genomics, where even a single true positive, correctly identifying that an individual's sequence was used in training,  constitutes a meaningful privacy breach given the immutability and identifiability of genomic data. Surveys by Hu et al.~\cite{hu2022mia_survey} and Wu and Cao~\cite{wu2025mia_largescale} further catalog the landscape of threshold-based, classifier-based, and metric-based MIA approaches across classification and generative model architectures.

For our setting, Duan et al.~\cite{duan2024mia_llm} conducted a large-scale evaluation of MIA on large language models ranging from 160M to 12B parameters and found that attacks barely outperform random guessing when training corpora are large. However, they identified that fine-tuned models trained on smaller, domain-specific datasets remain vulnerable, particularly when member and non-member distributions diverge. This finding directly motivates our experimental setup: GLMs fine-tuned on small genomic cohorts occupy precisely the vulnerability regime that Duan et al. identified, yet this configuration has not been evaluated in the genomic domain.

Beyond natural language, only Chen et al.~\cite{chen2021dp_genomic} evaluate membership inference on genomic machine learning, targeting Lasso and CNN models for phenotype prediction. Nemecek et al.~\cite{nemecek2025exploring} examined MIA on a clinical LLM fine-tuned on electronic health records, finding limited but measurable leakage; however, their work targets clinical text with a single attack category. The statistical properties of genomic data (i.e., high inter-individual correlation, familial linkage, and linkage disequilibrium) may alter MIA effectiveness, requiring dedicated evaluation for GLMs.

\subsection{Privacy Risks and Defenses for Genomic Data}\label{sec:genomic-privacy-rw}
Memorization of genomic sequences poses uniquely severe privacy risks compared to memorization of natural language text, owing to three distinctive properties of genomic data. First, a compromised genome is permanent, unlike credentials or financial data, it cannot be changed, rotated, or reissued~\cite{erlich2014routes, gymrek2013identifying}. Even trace genomic contributions to complex mixtures are identifiable: Homer et al.~\cite{homer2008resolving} showed that as little as 0.1\% of an individual's DNA in a pooled sample could be detected through SNP arrays and summary statistics, a result that led the NIH to restrict public access to aggregate GWAS data. Second, direct-to-consumer genetic databases have amplified this risk; Erlich et al.~\cite{erlich2018identity} showed that genealogy databases covering approximately 2\% of a target population suffice to identify nearly any individual through third-cousin or closer matches. In practice, cross-referencing a few hundred SNPs against consumer genealogy databases is sufficient to uniquely identify most individuals~\cite{erlich2014routes}. Third, genomic information is inherited, so a memorized sequence from one individual may expose sensitive variants shared by parents, siblings, or children who had no involvement in data collection~\cite{humbert2013kin}. This familial propagation of privacy risk has no direct analog in text-based memorization. The convergence of these three properties: immutability, identifiability from partial data, and heritability, means that even limited memorization in a GLM can produce cascading privacy harms that are neither reversible nor confined to the individual whose data was memorized.

\textit{Risk Landscape.}
Erlich and Narayanan~\cite{erlich2014routes} provided the definitive taxonomy of genetic privacy vulnerabilities, mapping three principal routes to breach: identity tracing, attribute disclosure, and completion attacks. Subsequent work has expanded this attack surface along several dimensions. Almadhoun et al.~\cite{almadhoun2020genomic_dp} demonstrated that inherent correlations in genomic data, such as familial relatedness and linkage disequilibrium, degrade the guarantees of differential privacy mechanisms designed for independent records. Ayoz et al.~\cite{ayoz2021genome_reconstruction} developed genome reconstruction attacks on genomic data-sharing beacons, showing that sequence-level information can be recovered from ostensibly privacy-preserving query interfaces. A recent systematic assessment~\cite{jmir2024genetic_privacy} further cataloged privacy vulnerabilities across diverse genetic data types and sharing modalities.

The most directly relevant prior work is that of Belfiore et al.~\cite{belfiore2025bihmia}, who introduced biHMIA, a biologically-informed hybrid membership inference attack that combines traditional black-box MIA with contextual genomics metrics, including mutation-specific features. Evaluated on GPT-like transformer architectures trained on genomic mutation profiles, biHMIA demonstrated that hybrid attacks incorporating biological domain knowledge outperform purely metric-based MIAs. The authors additionally evaluated differential privacy (DP) as a defense, finding that it reduced but did not eliminate attack success.

\textit{Defenses.}
In the genomic domain specifically, Chen et al.~\cite{chen2021dp_genomic} applied DP as a defense against membership inference on Lasso and CNN models for phenotype prediction, finding that DP is effective but imposes measurable utility degradation. Kolobkov et al.~\cite{kolobkov2024federated_genomic} explored an orthogonal approach through federated learning for genomic data using UK Biobank and 1000 Genomes datasets, distributing computation across sites to avoid centralized data aggregation. Training data deduplication, shown to reduce memorization and extraction success in the NLP setting~\cite{lee2022dedup, kandpal2022dedup_privacy}, represents a further mitigation strategy, though its effectiveness has not been evaluated on genomic training corpora.

However, the works surveyed above address privacy in traditional genomic settings, such as beacons, GWAS summary statistics, biobank databases, or through attacks on classifiers and generative models, not GLMs trained on raw nucleotide sequences. Our work differs from the most related prior effort, biHMIA~\cite{belfiore2025bihmia}, along three dimensions: data modality (raw nucleotide sequences rather than mutation profiles), attack breadth (three complementary vectors unified into a risk score rather than a single attack category), and controlled quantification (planted canary experiments measuring memorization as a function of repetition frequency).

\section{Threat Model}
\label{sec:threat-model}

We define the threat model under which our risk quantification framework operates, specifying the adversary's capabilities, objectives, and the assumptions under which a successful attack constitutes a meaningful privacy breach. 

\subsection{Data Sensitivity Assumption}\label{data-assumption}
We assume that the training corpus $\mathcal{D}_{\text{train}}$ may contain genomic sequences that are individually identifiable or that encode sensitive information about the individuals from whom they were derived. As established in Section~\ref{sec:genomic-privacy-rw}, even partial genomic sequences can be traced to specific individuals through re-identification attacks on external databases~\cite{homer2008resolving}, and genomic data is immutable once compromised~\cite{erlich2014routes}. Under this assumption, any successful recovery of training sequences or inference of training set membership constitutes a privacy breach with potentially irreversible consequences that may extend to biological relatives of the data subjects~\cite{humbert2013kin}.

\subsection{Adversarial Model}\label{adversary-model}
We consider an adversary $\mathcal{A}$ who has query access to a trained GLM $f_\theta$, where $\theta$ denotes the learned parameters. Depending on the specific attack, $\mathcal{A}$ may operate under white-box access (full access to model weights and architecture) or black-box access (ability to submit input sequences and observe output probabilities or generated tokens only). In both settings, the adversary interacts with the model solely at inference time.

\subsection{Objectives}\label{objectives}
The adversary pursues three goals, each corresponding to a component of our framework:

(1) \textbf{Memorization detection.} Determine whether the model exhibits differential behavior on sequences drawn from its training corpus versus sequences it has not observed during training. A model that assigns systematically higher confidence to training sequences than to held-out sequences of comparable genomic composition reveals a detectable memorization signal.

(2) \textbf{Sequence recovery.} Recover sequences (whole or partially) from the training corpus by exploiting the model's learned distribution. Success demonstrates that individual training examples are encoded within the model's parameters in a retrievable form.

(3) \textbf{Membership inference.} Given a candidate genomic sequence $x$, determine whether $x \in \mathcal{D}_{\text{train}}$. This binary classification task quantifies whether the model's behavior on $x$ reveals information about training set membership. 

These three objectives are not fully independent as successful sequence recovery implies both detectable memorization and inferable membership, and reliable membership inference presupposes differential model behavior on training versus held-out data. However, the converse relationships do not hold, and because each vector can succeed where the others fail, together they provide a multi-dimensional characterization of memorization vulnerability.

\subsection{Scope and Exclusions}\label{scope-exclusions}
We explicitly exclude adversaries who can modify the training procedure (e.g., data poisoning) or who have access to model gradients during training (e.g., gradient leakage attacks). Our threat model assumes the model has already been trained and is being evaluated or deployed. We additionally note that attribute inference attacks~\cite{fredrikson2015model_inversion, mehnaz2022attribute_inference}, which seek to infer sensitive properties of training data from model outputs, represent an important complementary threat. While attribute inference is beyond the scope of the present study, the modular design of our framework accommodates its future integration as an additional diagnostic component.

The complementary perspective to that of the adversary is that of the \emph{defender}: a model developer who seeks to quantify memorization risks prior to releasing a GLM, and who may apply the same risk quantification framework as an auditing tool to assess whether training data can be extracted or inferred from the model.

\section{Methodology}\label{Methods}
We first discuss the model configurations, and then the datasets, as well as the risk quantification protocol. 

\subsection{Model Configurations}\label{model-config}
To evaluate memorization risks across the current landscape of GLM architectures, we select four models that collectively span the principal design paradigms, parameter scales, and fine-tuning strategies represented in the field. Table~\ref{tab:models} summarizes their architectural characteristics.

\textbf{SimpleDNALM} is a standard four-layer causal transformer language model constructed for this study to serve as a controlled baseline. It is not a separately released or pretrained foundation model, but a minimal architecture consisting of token and positional embeddings, a transformer encoder with a hidden dimension of 512 and eight attention heads, followed by a linear language modeling head trained with next-token cross-entropy loss. With approximately 12.9M parameters and a vocabulary of eight tokens (four nucleotides plus special tokens), SimpleDNALM provides a setting in which memorization dynamics are isolated from the complexity of large-scale pretraining.

\textbf{DNABERT-2}~\cite{zhou2024dnabert2} represents the masked language modeling paradigm, pretrained on multi-species genomes with byte-pair encoding tokenization. At 117M parameters, it is the most widely adopted GLM for downstream genomic tasks.

\textbf{HyenaDNA}~\cite{nguyen2023hyenadna} employs a subquadratic long-range convolutional architecture that operates at single-nucleotide resolution. At 14.2M parameters, it represents the autoregressive paradigm with an implicit long-range dependency mechanism distinct from standard attention.

\textbf{Evo}~\cite{nguyen2024evo} is a 7B-parameter model based on the StripedHyena architecture, trained autoregressively on prokaryotic and eukaryotic genomes. Its substantially larger parameter count positions it at the frontier of GLM scale, making it a critical test case for whether increased model capacity amplifies memorization risk.

SimpleDNALM, DNABERT-2, and HyenaDNA are fine-tuned with full parameter updates. Evo is fine-tuned using Low-Rank Adaptation (LoRA)~\cite{hu2022lora}. This distinction introduces fine-tuning strategy as an additional experimental variable as full fine-tuning updates all parameters and may facilitate memorization through greater effective capacity, whereas LoRA constrains the update to a low-rank subspace, potentially limiting the model's ability to encode individual training examples. All models are trained on NVIDIA A100 GPUs with early stopping, and each configuration is repeated across three seeds to estimate variance. Training hyperparameters are reported in the supplementary material (Appendix~A).

\begin{table}[t]
\centering
\footnotesize
\caption{Models evaluated. FT = fine-tuning strategy.}
\label{tab:models}
\begin{tabular}{lllll}
\toprule
\textbf{Model} & \textbf{Architecture} & \textbf{Params} & \textbf{FT} & \textbf{Tokenization} \\
\midrule
SimpleDNALM & Causal Transformer & 12.9M & Full & Character-level \\
DNABERT-2   & BERT Encoder        & 117M            & Full & BPE \\
HyenaDNA    & Hyena (Conv)        & 14.2M           & Full & Single-nucleotide \\
Evo         & StripedHyena (SSM)  & 7B              & LoRA & Byte-level \\
\bottomrule
\end{tabular}
\end{table}

\subsection{Datasets}\label{datasets}
We evaluate each model on four genomic datasets that vary in biological complexity, ranging from sequences with no biological structure to real eukaryotic genomes. Table~\ref{tab:datasets} summarizes the datasets used.

\textbf{Synthetic} sequences are generated as independent, identically distributed nucleotides with a balanced GC content of 0.5, producing sequences with no motifs, codon structure, or evolutionary conservation. This zero-order model provides a controlled baseline in which any detected memorization cannot be attributed to biological regularity.

\textbf{E.\ coli} sequences are from the \textit{Escherichia coli} K-12 MG1655 reference genome~\cite{blattner1997complete} (RefSeq accession GCF\_000005845.2~\cite{NCBI_GCF_000005845_2}) using a sliding window with a stride of 256 nucleotides. As a well-characterized prokaryotic genome of approximately 4.6 Mb, E.\ coli provides gene-dense sequences with moderate repetitive content.

\textbf{Yeast} sequences are from the \textit{Saccharomyces cerevisiae} S288C~\cite{goffeau1996life, engel2014reference} reference genome (RefSeq accession GCF\_000146045.2~\cite{NCBI_GCF_000146045_2}) using the same windowing procedure. At approximately 12 Mb across 16 chromosomes, the yeast genome introduces eukaryotic complexity including intergenic regions, transposable elements, and greater structural variation relative to E. coli.

\textbf{GUE} sequences are from the Genomic Understanding Evaluation benchmark~\cite{zhou2024dnabert2}, specifically the \texttt{prom\_300\_all} promoter prediction subset accessed via HuggingFace. Unlike the reference genome datasets, GUE provides curated, functionally annotated sequences from a multi-species collection.

Across all four datasets, we use 1{,}000 training sequences and 200 test sequences of length 256 nucleotides. This corpus size is deliberately small, reflecting the regime in which GLMs are fine-tuned on limited domain-specific cohorts, the setting identified by Duan et al.~\cite{duan2024mia_llm} as most vulnerable to memorization. For the real genomic sources, sequences containing ambiguous bases are discarded and remaining sequences are center-cropped to the target length. Train and test splits are drawn by random sampling and shuffling, seeded for reproducibility.

\subsection{Risk Quantification Protocol}\label{risk-quantification-protocol}
We instantiate three adversarial objectives (see Section~\ref{sec:threat-model}) as three complementary evaluation vectors unified through a maximum vulnerability score. Central to this evaluation is the use of canary sequences, synthetic probes planted in the training corpus to enable controlled measurement of memorization dynamics. Evaluation vectors are executed on the best checkpoint selected by lowest loss on the held-out test split, which also serves as the early stopping criterion; this is conservative, as selecting the checkpoint that minimizes test loss compresses the train-test gap exploited by the perplexity and membership inference vectors, meaning a dedicated validation split would likely yield higher vulnerability scores.

\subsubsection{Canary Design}\label{canary-design}
Following the canary methodology introduced by Carlini et al.~\cite{carlini2019secret}, we generate 100 distinct canary sequences of 64 nucleotides each, sampled independently and identically from a uniform nucleotide distribution with 50\% GC content. Canaries carry no biological structure, motifs, or marker patterns; they are indistinguishable in composition from the synthetic baseline dataset, differing from real genomic sequences only in the absence of evolutionary or functional constraints. This design ensures that any preferential treatment by the model, such as anomalously low perplexity or successful extraction, can be attributed to memorization rather than biological plausibility.

Each canary is assigned to one of four repetition tiers: 1, 5, 10, or 20 duplicate copies inserted into the training corpus, with 25 canaries per tier. Insertion positions within the training sequence list are selected uniformly at random. This tiered design enables direct measurement of how data duplication drives memorization, the central scaling relationship. Canaries are inserted only into the training set and are never included in the test split. The canary generation seed is derived deterministically from the experiment seed to ensure reproducibility across runs. To confirm that canaries cannot be identified and filtered by a practitioner through inspection of the training corpus, we conduct a canary detectability analysis (see the supplementary material, Appendix~C).

\subsubsection{Canary Extraction} This evaluation vector targets the sequence recovery objective. For each canary, we condition the model on a prefix of the canary sequence and attempt to recover the remainder via beam search with a beam width of 10 and a maximum of 1{,}000 candidate sequences. The true canary is then ranked among all candidates by model likelihood, and recovery is quantified using the exposure metric introduced by Carlini et al.~\cite{carlini2019secret}. For a canary sequence $s$ inserted into the training corpus, exposure is defined as:
\begin{equation}
\text{Exposure}(s) = \log_2 |R| - \log_2(\text{rank}(s)),
\label{eq:exposure}
\end{equation}
where $|R|$ denotes the size of the randomness space from which $s$ was drawn and $\text{rank}(s)$ is the rank of $s$ among all possible sequences in $R$ when ordered by decreasing model likelihood. An exposure equal to $\log_2 |R|$ indicates that the model assigns the highest likelihood to the true canary, representing complete memorization. We report extraction success rates and mean exposure scores disaggregated by repetition tier.

\begin{table}[t]
\centering
\footnotesize
\caption{Datasets evaluated across all model types.}
\label{tab:datasets}
\begin{tabular}{llll}
\toprule
\textbf{Dataset} & \textbf{Source} & \textbf{Organism} & \textbf{Biological Structure} \\
\midrule
Synthetic & Generated & --- & None (zero-order) \\
E.\ coli  & NCBI RefSeq & \textit{E.\ coli} K-12 & Prokaryotic \\
Yeast     & NCBI RefSeq & \textit{S.\ cerevisiae} S288C & Eukaryotic \\
GUE       & HuggingFace & Multi-species &  Promoter regions \\
\bottomrule
\end{tabular}
\end{table}

\subsubsection{Perplexity-Based Detection} This evaluation vector targets the memorization detection objective by exploiting the observation that memorized sequences receive disproportionately low loss under the trained model. We compute per-sequence perplexity across three disjoint splits: training sequences (excluding canaries), held-out test sequences, and canary sequences. A model exhibiting memorization will assign systematically lower perplexity to training and canary sequences relative to test sequences of comparable composition. We additionally disaggregate canary perplexity by repetition tier, enabling analysis of how insertion frequency modulates the model's confidence on memorized sequences. The gap between canary and test perplexity distributions serves as the primary diagnostic signal for this vector.

\subsubsection{Membership Inference} This evaluation vector targets the membership inference objective. We employ a likelihood-ratio attack that fits parametric distributions to the training and test loss populations and classifies membership based on the resulting likelihood ratio. We report the area under the receiver operating characteristic curve (AUC-ROC) as the primary metric, where a value of 0.5 indicates no membership leakage and 1.0 indicates perfect discrimination.

\begin{table*}[t]
\centering
\small
\setlength{\tabcolsep}{3.5pt}
\caption{Per-vector memorization risk scores and maximum vulnerability scores for each model-dataset configuration. Component scores ($s_{\text{ppl}}, s_{\text{ext}}, s_{\text{mia}}$) are reported as average across seeds. $S_{\text{config}}$ and $S_{\text{model}}$ are computed per seed as the maximum of the component scores for that seed, then reported as mean $\pm$ std across seeds. Bold indicates the maximum vector driving $S_{\text{config}}$.}
\label{tab:vulnerability-scores}
\begin{tabular}{@{}l cccc cccc cccc cccc c@{}}
\toprule
& \multicolumn{4}{c}{\textbf{Synthetic}}
& \multicolumn{4}{c}{\textbf{E.\ coli}}
& \multicolumn{4}{c}{\textbf{Yeast}}
& \multicolumn{4}{c}{\textbf{GUE}}
& \\
\cmidrule(lr){2-5} \cmidrule(lr){6-9} \cmidrule(lr){10-13} \cmidrule(lr){14-17}
\textbf{Model}
& $s_{\text{ppl}}$ & $s_{\text{ext}}$ & $s_{\text{mia}}$ & $S_{\text{config}}$
& $s_{\text{ppl}}$ & $s_{\text{ext}}$ & $s_{\text{mia}}$ & $S_{\text{config}}$
& $s_{\text{ppl}}$ & $s_{\text{ext}}$ & $s_{\text{mia}}$ & $S_{\text{config}}$
& $s_{\text{ppl}}$ & $s_{\text{ext}}$ & $s_{\text{mia}}$ & $S_{\text{config}}$
& $S_{\text{model}}$ \\
\midrule

SimpleDNALM
& .02 & .50 & \textbf{.52} & \riskcell{0.53}{.53{\scriptsize$\pm$.02}}
& .01 & .45 & \textbf{.47} & \riskcell{0.48}{.48{\scriptsize$\pm$.00}}
& .00 & .48 & \textbf{.49} & \riskcell{0.50}{.50{\scriptsize$\pm$.02}}
& .00 & \textbf{.51} & .45 & \riskcell{0.51}{.51{\scriptsize$\pm$.04}}
& \riskcell{0.55}{.55{\scriptsize$\pm$.01}} \\

DNABERT-2
& .35 & .14 & \textbf{.48} & \riskcell{0.48}{.48{\scriptsize$\pm$.02}}
& .34 & .14 & \textbf{.43} & \riskcell{0.43}{.43{\scriptsize$\pm$.01}}
& .38 & .17 & \textbf{.45} & \riskcell{0.45}{.45{\scriptsize$\pm$.02}}
& .38 & .09 & \textbf{.45} & \riskcell{0.45}{.45{\scriptsize$\pm$.01}}
& \riskcell{0.48}{.48{\scriptsize$\pm$.01}} \\

HyenaDNA
& -.36 & .23 & \textbf{.48} & \riskcell{0.48}{.48{\scriptsize$\pm$.01}}
& -1.04 & .15 & \textbf{.49} & \riskcell{0.49}{.49{\scriptsize$\pm$.01}}
& -.05 & .15 & \textbf{.49} & \riskcell{0.49}{.49{\scriptsize$\pm$.01}}
& -.90 & .21 & \textbf{.47} & \riskcell{0.47}{.47{\scriptsize$\pm$.01}}
& \riskcell{0.49}{.49{\scriptsize$\pm$.00}} \\

Evo (LoRA)
& .20 & \textbf{.67} & .48 & \riskcell{0.82}{.82{\scriptsize$\pm$.23}}
& .34 & \textbf{1.00} & .57 & \riskcell{1.00}{1.00{\scriptsize$\pm$.00}}
& .33 & \textbf{1.00} & .50 & \riskcell{1.00}{1.00{\scriptsize$\pm$.00}}
& .34 & \textbf{1.00} & .40 & \riskcell{1.00}{1.00{\scriptsize$\pm$.00}}
& \riskcell{1.00}{1.00{\scriptsize$\pm$.00}} \\

\bottomrule
\end{tabular}
\end{table*}

\subsubsection{Maximum Vulnerability Score}
The outputs of all three evaluation vectors are unified into a single memorization risk score in $[0, 1]$ that provides a standardized summary of memorization vulnerability. Each component is first normalized to $[0, 1]$ with higher values indicating greater risk. 

The perplexity score is computed as $s_{\text{ppl}}(\theta, d) = 1 - \bar{p}_{\text{canary}} / \bar{p}_{\text{test}}$, where $\bar{p}_{\text{canary}}$ and $\bar{p}_{\text{test}}$ are the mean perplexities on the canary and held-out test splits, respectively, for model $\theta$ evaluated on dataset $d$. A value near 0 indicates the model treats canary sequences comparably to unseen data with positive values indicating memorization (the model assigns lower perplexity to canaries than to held-out data), and negative values indicating the model finds canaries harder than unseen sequences. The extraction score $s_{\text{ext}}(\theta, d) \in [0, 1]$ is the fraction of canary sequences successfully recovered via beam search, directly quantifying whether memorized content is retrievable by an adversary. The membership inference score is computed as $s_{\text{mia}}(\theta, d) = \text{max}(0, 2\cdot(\text{AUC}-0.5))$, where AUC is the area under the ROC curve of the likelihood ratio attack. This rescaling maps the chance-level baseline of 0.5 to a score of 0, ensuring that $s_{\text{mia}}$ shares a common zero point with the other two components.

We adopt a worst-case formulation rather than a weighted average, following established practice in vulnerability testing and adversarial evaluation frameworks~\cite{nist80030r1}. In such regimes, a system is assessed by running a battery of tests spanning different attack surfaces, and the system's risk is determined by its worst performance under any single test. The rationale is that a model's privacy exposure is governed by its most exploitable weakness: if any single attack vector reveals memorization, the model is vulnerable regardless of how it performs on the others. This logic extends naturally across evaluation datasets as well. Because the true data distribution a deployed model will encounter is unknown, we evaluate across datasets spanning different biological structures and complexities, and characterize risk by the worst-case outcome, providing a conservative bound on the model's maximum vulnerability. For a given model $\theta$ and dataset $d$, we define:
\begin{equation}
S_{\text{config}}(\theta, d) = \max(s_{\text{ppl}}(\theta, d),\; s_{\text{ext}}(\theta, d),\; s_{\text{mia}}(\theta, d)),
\label{eq:sconfig}
\end{equation}
and for a given model across all evaluation datasets:
\begin{equation}
S_{\text{model}}(\theta) = \max_{d \in \mathcal{D}}\; S_{\text{config}}(\theta, d),
\label{eq:smodel}
\end{equation}
where $\mathcal{D}$ denotes the set of datasets. $S_\text{config}(\theta, d)$ captures the maximum vulnerability of a specific model-dataset pair across all attack vectors, while $S_\text{model}(\theta)$ answers the auditing question: what is the maximum vulnerability of this model under any data regime evaluated? All composite scores are computed per seed before averaging; that is, the reported $S_\text{config}$ is the mean over seeds of the per-seed maximum, and the reported $S_\text{model}$ is the mean over seeds of the per-seed maximum $S_\text{config}$ across datasets. We report both the maximum vulnerability scores and their constituent components to support fine-grained interpretation of where memorization risk is concentrated and which attack vector drives it.

\section{Results}\label{Results}
We evaluate all four models across four datasets and three seeds. For each configuration, we report the maximum vulnerability score and its constituent components: canary extraction success, perplexity-based detection, and membership inference.

\subsection{Maximum Vulnerability Scores}\label{mem-risk}
Table~\ref{tab:vulnerability-scores} presents the per-vector memorization risk scores and maximum vulnerability scores for each model-dataset configuration. The maximum vulnerability score $S_{\text{config}}$ captures the worst-case risk across all three attack vectors for a given configuration, while $S_{\text{model}}$ reports the worst-case risk for each model across all datasets evaluated.

The results reveal a sharp separation between Evo (LoRA) and the remaining architectures. Evo achieves $S_{\text{model}} = 1.00 \pm 0.00$, driven entirely by canary extraction ($s_{\text{ext}} = 1.00$) on all three real genomic datasets. On synthetic data, Evo's vulnerability drops substantially ($S_{\text{config}} = 0.82 \pm 0.23$) with high variance across seeds, suggesting that the absence of biological structure removes contextual regularities that facilitate extraction in a large pretrained model. This result is notable given that Evo is the only model fine-tuned with LoRA rather than full parameter updates, indicating that parameter-efficient fine-tuning of a larger model does not inherently reduce memorization risk relative to full fine-tuning of smaller models.

The remaining three models cluster within a narrower band ($S_{\text{model}}$ between 0.48 and 0.55), but their vulnerability profiles differ in a way that underscores the necessity of multi-vector evaluation. DNABERT-2 and HyenaDNA are both dominated by membership inference across all four datasets, with $s_{\text{mia}}$ consistently exceeding $s_{\text{ppl}}$ and $s_{\text{ext}}$. However, DNABERT-2 maintains the highest perplexity scores of any model ($s_{\text{ppl}} = 0.34$-$0.38$) alongside the lowest extraction scores ($s_{\text{ext}} = 0.09$-$0.14$), indicating that its memorization is encoded in a form detectable through loss-based analysis but not recoverable through sequential generation. HyenaDNA, by contrast, exhibits near-zero perplexity signal and low extraction, leaving membership inference as its only measurable vulnerability. SimpleDNALM is the only model whose dominant vector shifts across datasets: $s_{\text{mia}}$ drives $S_{\text{config}}$ on Synthetic, E.\ coli, and Yeast, while $s_{\text{ext}}$ takes over on GUE ($0.51 \pm 0.04$), reflecting a more balanced vulnerability profile in which extraction and membership inference contribute comparably. Per-seed $S_{\text{config}}$ distributions, $S_{\text{model}}$, and the individual component scores ($s_{\text{ppl}}, s_{\text{ext}}, s_{\text{mia}}$) used to compute them are reported in the supplementary material (Appendix~B), confirming that the findings are stable across training runs and enabling independent verification of the aggregation.

\textbf{Key Finding:} Evo (LoRA) separates sharply from all other architectures with $S_{\text{model}} = 1.00$, demonstrating that parameter-efficient fine-tuning alone is insufficient to limit memorization in a large model. The remaining models cluster between 0.48 and 0.55 but are dominated by different attack vectors, confirming that multi-vector evaluation is necessary to capture the full memorization risk profile.

\subsection{Canary Extraction and Exposure}\label{sec:extraction}
\begin{figure}[t]
\centering
\begin{subfigure}[b]{0.49\columnwidth}
    \centering
    \includegraphics[width=\textwidth]{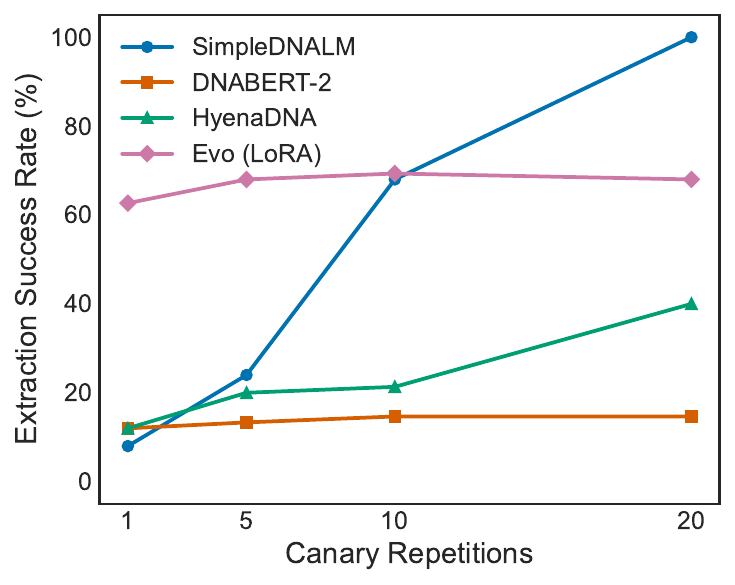}
    \caption{Synthetic}
    \label{fig:ext_synthetic}
\end{subfigure}
\hfill
\begin{subfigure}[b]{0.49\columnwidth}
    \centering
    \includegraphics[width=\textwidth]{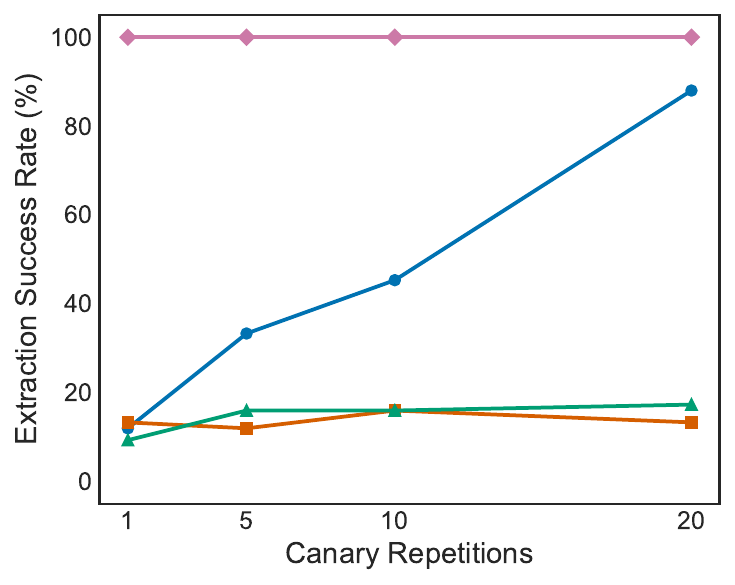}
    \caption{E.\ coli}
    \label{fig:ext_ecoli}
\end{subfigure}
\\[4pt]
\begin{subfigure}[b]{0.49\columnwidth}
    \centering
    \includegraphics[width=\textwidth]{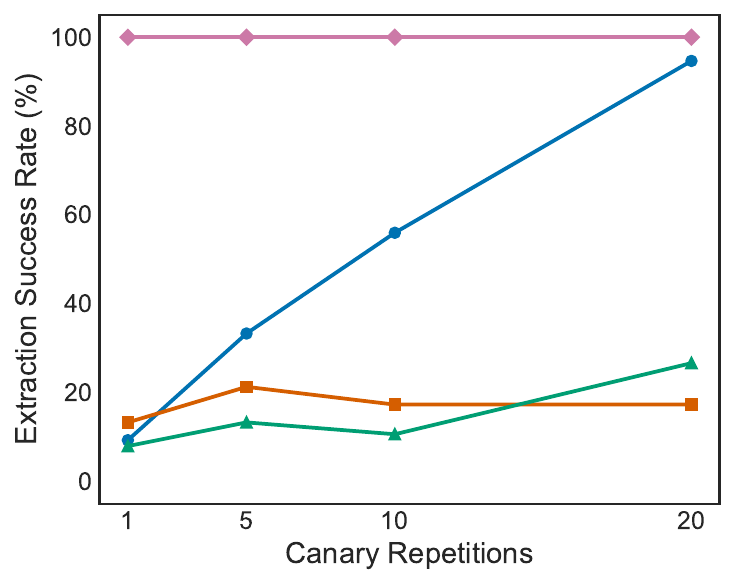}
    \caption{Yeast}
    \label{fig:ext_yeast}
\end{subfigure}
\hfill
\begin{subfigure}[b]{0.49\columnwidth}
    \centering
    \includegraphics[width=\textwidth]{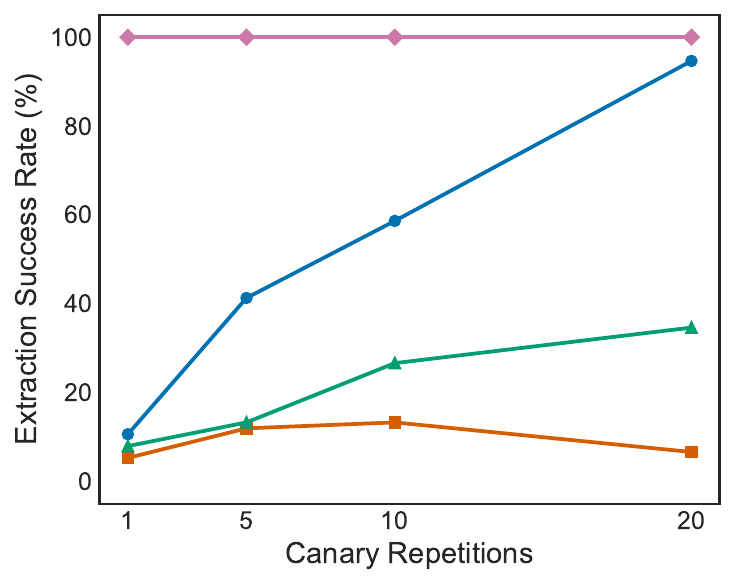}
    \caption{GUE}
    \label{fig:ext_gue}
\end{subfigure}
\caption{Canary extraction success rate as a function of repetition tier across datasets. Evo (LoRA) achieves near-complete extraction on real genomic data regardless of repetition count, while DNABERT-2 remains resistant.}
\label{fig:extraction}
\end{figure}

Figure~\ref{fig:extraction} presents canary extraction success rates as a function of repetition tier across all four datasets. The results reveal sharply divergent memorization profiles across architectures.

Evo (LoRA) suffers from near-complete extraction on all three real genomic datasets regardless of repetition count, recovering 100\% of canaries even at a single insertion on E.\ coli, Yeast, and GUE. On synthetic data, however, Evo's extraction begins lower (approximately 63\%) and plateaus near 69\%, suggesting that in the absence of biological structure the model's capacity advantage alone is insufficient for complete memorization. In contrast, DNABERT-2 demonstrates the strongest resistance to extraction, remaining approximately flat between 12-15\% across all repetition tiers and datasets with negligible sensitivity to data duplication. One hypothesis is that this resistance reflects the influence of DNABERT-2's masked language modeling pretraining objective, which may produce bidirectional representations that distribute sequence information across context rather than encoding the sequential patterns exploited by prefix-based extraction. SimpleDNALM exhibits the clearest monotonic scaling with repetition, rising steeply from approximately 8-12\% at a single insertion to 88-100\% at 20 insertions, directly confirming that the duplication-driven memorization scaling laws established by Carlini et al.~\cite{carlini2023quantifying} for natural language models extend to the genomic domain. HyenaDNA shows a more modest upward trend, reaching 18-40\% at 20 insertions depending on the dataset, consistent with a long-range convolutional architecture that partially resists rote memorization while remaining more susceptible than DNABERT-2.

Across all models, the dataset effect is secondary to the architectural effect: the relative ordering of models is preserved across synthetic, prokaryotic, eukaryotic, and curated benchmark data. This consistency suggests that the observed memorization profiles are driven primarily by model architecture and fine-tuning strategy rather than by properties of the training corpus.

\textbf{Key Finding:} Duplication-driven memorization scaling laws from natural language models transfer to the genomic domain, with SimpleDNALM showing clear monotonic scaling from 8-12\% to 88-100\% extraction as repetitions increase. Architecture is the primary determinant of extraction susceptibility, with model ordering preserved across all four datasets regardless of biological complexity.

\subsection{Memorization Leakage}
\subsubsection{Perplexity-Based Detection}
Table~\ref{tab:perplexity} reports mean perplexity on the training, test, and canary splits for each model-dataset configuration, along with the memorization gap ratio defined as test perplexity divided by canary perplexity. A gap ratio greater than 1.0 indicates that the model assigns lower perplexity to canary sequences than to held-out test data, revealing a detectable memorization signal.
\begin{table}[h]
\centering
\scriptsize
\caption{Mean perplexity across data splits and memorization gap ratio (test divided by canary perplexity) reported as average $\pm$ std across seeds. Gap $> 1.0$ indicates the model assigns lower perplexity to canaries than to test sequences, signaling memorization.}
\label{tab:perplexity}
\begin{tabular}{llcccc}
\toprule
\textbf{Model} & \textbf{Dataset} & \textbf{Train} & \textbf{Test} & \textbf{Canary} & \textbf{Gap} \\
\midrule
\multirow{4}{*}{SimpleDNALM}
  & Synthetic & 3.93 $\pm$ 0.00 & 4.00 $\pm$ 0.00 & 3.93 $\pm$ 0.01 & 1.02 $\pm$ 0.00 \\
  & E.\ coli  & 3.91 $\pm$ 0.00 & 3.94 $\pm$ 0.00 & 3.91 $\pm$ 0.01 & 1.01 $\pm$ 0.00 \\
  & Yeast     & 3.84 $\pm$ 0.02 & 3.86 $\pm$ 0.00 & 3.84 $\pm$ 0.02 & 1.00 $\pm$ 0.01 \\
  & GUE       & 3.82 $\pm$ 0.03 & 3.84 $\pm$ 0.01 & 3.83 $\pm$ 0.02 & 1.00 $\pm$ 0.00 \\
\midrule
\multirow{4}{*}{DNABERT-2}
  & Synthetic & 1.75 $\pm$ 0.00 & 2.69 $\pm$ 0.04 & 1.74 $\pm$ 0.00 & 1.54 $\pm$ 0.02 \\
  & E.\ coli  & 1.68 $\pm$ 0.03 & 2.51 $\pm$ 0.03 & 1.66 $\pm$ 0.01 & 1.51 $\pm$ 0.02 \\
  & Yeast     & 1.89 $\pm$ 0.01 & 3.03 $\pm$ 0.05 & 1.88 $\pm$ 0.03 & 1.61 $\pm$ 0.02 \\
  & GUE       & 1.79 $\pm$ 0.01 & 2.83 $\pm$ 0.06 & 1.77 $\pm$ 0.03 & 1.60 $\pm$ 0.01 \\
\midrule
\multirow{4}{*}{HyenaDNA}
  & Synthetic & 47.55 $\pm$ 13.69 & 35.76 $\pm$ 7.44 & 47.74 $\pm$ 14.15 & 0.80 $\pm$ 0.25 \\
  & E.\ coli  & 30.23 $\pm$ 11.25 & 14.95 $\pm$ 0.58 & 30.87 $\pm$ 11.80 & 0.55 $\pm$ 0.18 \\
  & Yeast     & 25.27 $\pm$ 9.91 & 23.59 $\pm$ 7.93 & 25.30 $\pm$ 9.99 & 0.96 $\pm$ 0.06 \\
  & GUE       & 61.45 $\pm$ 19.64 & 32.03 $\pm$ 6.77 & 62.29 $\pm$ 20.24 & 0.53 $\pm$ 0.06 \\
\midrule
\multirow{4}{*}{Evo (LoRA)}
  & Synthetic & 3.21 $\pm$ 0.57 & 4.00 $\pm$ 0.00 & 3.20 $\pm$ 0.57 & 1.29 $\pm$ 0.21 \\
  & E.\ coli  & 2.55 $\pm$ 0.03 & 3.86 $\pm$ 0.02 & 2.53 $\pm$ 0.04 & 1.53 $\pm$ 0.03 \\
  & Yeast     & 2.59 $\pm$ 0.06 & 3.83 $\pm$ 0.01 & 2.57 $\pm$ 0.07 & 1.50 $\pm$ 0.04 \\
  & GUE       & 2.57 $\pm$ 0.03 & 3.85 $\pm$ 0.03 & 2.55 $\pm$ 0.02 & 1.51 $\pm$ 0.00 \\
\bottomrule
\end{tabular}
\end{table}

DNABERT-2 exhibits the strongest perplexity-based memorization signal, with gap ratios between 1.51 and 1.61 across all datasets. Despite resisting canary extraction (Section~\ref{sec:extraction}), DNABERT-2 assigns lower perplexity to canary sequences (1.66-1.88) than to test sequences (2.51-3.03), indicating that memorization is encoded within the model's representations even when it is not recoverable through prefix-based generation. Evo (LoRA) shows a similar pattern, with gap ratios of 1.50-1.53 on real genomic datasets and a lower ratio of 1.29 on synthetic data, consistent with the reduced extraction performance on synthetic data observed in Figure~\ref{fig:extraction}.

SimpleDNALM presents a contrasting profile as its gap ratios hover near 1.0 (1.00-1.02) across all datasets, meaning canary and test perplexities are nearly indistinguishable. This is notable given that SimpleDNALM achieves high extraction success rates at elevated repetition tiers, suggesting that its memorization manifests through recoverable sequential patterns rather than through a broad reduction in loss on memorized sequences.

HyenaDNA operates in a substantially higher perplexity regime (14.95-62.29) than the other models, reflecting weaker overall fit to the training data. Despite this, its canary perplexity closely tracks its training perplexity across all datasets, with both consistently exceeding test perplexity. The resulting gap ratios fall below 1.0 (0.53-0.96), indicating that canary sequences are no easier for the model than held-out data and that perplexity-based detection would not identify memorization in HyenaDNA. This is consistent with its low vulnerability scores and modest extraction rates.

\textbf{Key Finding:} Memorization manifests differently across architectures at the perplexity level. DNABERT-2 shows the strongest perplexity gap (1.51-1.61) despite resisting extraction, while SimpleDNALM shows near-zero perplexity signal despite high extraction success, demonstrating that memorization is not a unitary phenomenon and that different vectors capture fundamentally different forms of information leakage.

\subsubsection{Membership Inference}
Table~\ref{tab:mia} reports the AUC-ROC of the likelihood ratio attack (LiRA) for each model-dataset configuration. Across all 16 configurations, AUC values range from 0.70 to 0.79, indicating that an adversary can infer training set membership at rates moderately above random guessing but without high reliability. Evo (LoRA) on E.\ coli achieves the highest AUC ($0.79 \pm 0.02$), consistent with its elevated extraction success and maximum vulnerability score on that dataset. Evo also produces the lowest AUC ($0.70 \pm 0.01$) on GUE, making it the most variable model across datasets. SimpleDNALM, DNABERT-2, and HyenaDNA exhibit narrower variation, clustering between 0.71 and 0.76. Notably, HyenaDNA's AUC (0.73-0.74) is comparable to the other models despite its substantially lower maximum vulnerability and extraction rates, suggesting that even models with limited memorization as measured by other vectors retain a detectable membership signal through loss-based inference.

\begin{table}[h]
\centering
\scriptsize
\caption{Membership inference AUC-ROC via likelihood ratio attack (mean $\pm$ std). AUC = 0.5 indicates random guessing; AUC = 1.0 indicates perfect membership discrimination.}
\label{tab:mia}
\begin{tabular}{lcccc}
\toprule
\textbf{Model} & \textbf{Synthetic} & \textbf{E.\ coli} & \textbf{Yeast} & \textbf{GUE} \\
\midrule
SimpleDNALM   & 0.76 $\pm$ 0.01 & 0.74 $\pm$ 0.00 & 0.75 $\pm$ 0.01 & 0.72 $\pm$ 0.00 \\
DNABERT-2     & 0.74 $\pm$ 0.01 & 0.71 $\pm$ 0.01 & 0.73 $\pm$ 0.01 & 0.72 $\pm$ 0.01 \\
HyenaDNA      & 0.74 $\pm$ 0.00 & 0.74 $\pm$ 0.00 & 0.74 $\pm$ 0.00 & 0.73 $\pm$ 0.01 \\
Evo (LoRA)    & 0.74 $\pm$ 0.01 & 0.79 $\pm$ 0.02 & 0.75 $\pm$ 0.00 & 0.70 $\pm$ 0.01 \\
\bottomrule
\end{tabular}
\end{table}

\textbf{Key Finding:} All models exhibit membership inference AUC between 0.70 and 0.79, confirming that a detectable membership signal persists across all architectures regardless of their extraction or perplexity profiles. Even HyenaDNA, which shows minimal vulnerability on other vectors, retains comparable membership leakage through loss-based inference.

\section{Discussion and Limitations}\label{Discussion}
\textbf{Main findings.} Our results establish that the memorization scaling laws identified in natural language models by Carlini et al.~\cite{carlini2023quantifying} transfer to the genomic domain. The canary extraction experiments provide the most direct evidence: SimpleDNALM exhibits a clear monotonic relationship between data duplication and extraction success (Figure~\ref{fig:extraction}), and HyenaDNA follows the same directional trend at lower absolute rates. These results confirm that duplication-driven memorization is a general property of sequence models trained with gradient-based optimization on repeated examples, not an artifact of natural language structure or large vocabulary spaces. A second consistent finding is that model architecture is the primary determinant of memorization profile, with the relative ordering of models preserved across all four datasets regardless of biological complexity. This suggests that architectural inductive biases and fine-tuning strategy govern memorization dynamics more than the statistical properties of the training corpus.

Most importantly, the three evaluation vectors capture qualitatively distinct dimensions of memorization risk, and their joint analysis reveals privacy profiles that any single-metric evaluation would miss. DNABERT-2 illustrates this most clearly where the model demonstrates the strongest resistance to prefix-based canary extraction (12-15\% across all repetition tiers and datasets) yet produces the highest perplexity gap ratios (1.51-1.61). This result indicates that memorized information is encoded within the model's representations in a form that is detectable through loss-based analysis but not recoverable through sequential generation. Our hypothesis is that this dissociation reflects the masked language modeling objective, which may produce bidirectional representations that distribute sequence information across context rather than encoding the left-to-right dependencies exploited by prefix-conditioned extraction. However, the methodology does not isolate this factor from other confounding variables, including parameter count and tokenization strategy, and a controlled ablation comparing pretraining objectives on a fixed architecture would be needed to establish a causal relationship. SimpleDNALM presents the inverse with high extraction success at elevated repetition tiers paired with gap ratios near 1.0 (1.00-1.02), indicating that memorization manifests as recoverable sequential patterns without producing a reduction in loss relative to held-out data. These contrasts demonstrate that memorization is not a unitary phenomenon.

The behavior of Evo (LoRA) challenges the assumption that parameter-efficient fine-tuning inherently limits memorization. Despite updating only a low-rank subspace of its 7B parameters, Evo achieves the highest maximum vulnerability scores on three of four datasets and recovers 100\% of canary sequences on all real genomic datasets regardless of repetition count. One plausible explanation is that the pretrained parameters already encode sufficient representational capacity and distributional knowledge of nucleotide sequences that LoRA fine-tuning on a small corpus concentrates parameter updates on memorizing the specific training examples rather than learning generalizable patterns. This hypothesis is consistent with the observation that Evo's extraction success drops on synthetic data (63-69\%), where the absence of biological structure removes the contextual regularities that the pretrained representations may exploit. Put differently, biological structure in real genomic sequences may provide the model with informative prefixes and contextual cues that facilitate extraction, whereas random sequences offer no such leverage. This interpretation rests on a single model at a single LoRA rank, and further investigation across LoRA configurations and model scales would be needed.

\textbf{Practical implications.} The maximum vulnerability scores reveal that memorization risk varies across architectures, with Evo (LoRA) reaching $S_{\text{model}} = 1.00$ while the remaining models cluster between 0.48 and 0.55. Even among the lower-scoring models, these scores are driven by at least one attack vector exceeding moderate levels, confirming that under standard fine-tuning conditions, every architecture evaluated retains a detectable memorization signal. For practitioners operating under regulatory frameworks~\cite{naveed2015privacy_genomic, bonomi2020genomic_sharing}, these findings suggest that releasing fine-tuned GLMs without privacy auditing constitutes a non-trivial compliance risk. The component-level decomposition underlying each maximum vulnerability score is particularly relevant as a model may appear safe under one evaluation criterion while remaining vulnerable under another. We recommend that privacy risk assessments for GLMs adopt multi-vector evaluation as a minimum standard, and that maximum vulnerability scores be interpreted alongside their constituent components to identify which attack vector drives the risk for a given model.

\textbf{Limitations and future work.} SimpleDNALM, while valuable as a controlled baseline for isolating memorization dynamics, is a lightweight transformer and does not reflect the pretraining regimes of production GLMs. Conclusions drawn from its behavior may not generalize to models deployed in clinical pipelines. While canary length is evaluated as a scaling dimension (supplementary material, Appendix~D), the canaries remain random nucleotide strings carrying no biological structure. While this design ensures that detected memorization is attributable to training dynamics rather than biological plausibility, it remains an open question whether memorization of real sequences would manifest at comparable rates or through the same vectors. Additionally, our evaluation of Evo employed LoRA fine-tuning exclusively; we therefore cannot determine whether full parameter fine-tuning of a 7B-parameter model would change the memorization patterns observed relative to the smaller fully fine-tuned architectures. More broadly, all experiments use a fixed corpus size of 1,000 training sequences, and memorization dynamics may shift as dataset scale increases toward the cohort sizes typical of biobank-derived fine-tuning.

These limitations suggest directions for future work. Extending the framework to evaluate memorization under full fine-tuning of large-scale models, across a range of LoRA ranks and other parameter-efficient strategies, would clarify the relationship between fine-tuning capacity and memorization risk. Replacing synthetic canaries with biologically realistic probes, such as sequences carrying known pathogenic variants or promoter regions, would better approximate the real-world threat surface; the design space here is broad and requires careful calibration to ensure canaries are realistic enough to test clinically meaningful threats while remaining distinguishable for controlled evaluation. Evaluating memorization as a function of training corpus size, spanning from the small cohorts evaluated here through biobank-scale corpora, would establish whether the vulnerability scores observed at 1,000 sequences persist, diminish, or escalate under more realistic data regimes, providing directly actionable guidance for regulatory compliance.

Integrating privacy defenses into the framework is a natural next step. Differential privacy (DP) provides formal guarantees against memorization~\cite{carlini2019secret}, and evaluating how DP-trained models perform under the multi-vector attack surface characterized here would directly connect our auditing framework to defense efficacy. The membership inference component would also benefit from alternative privacy metrics such as bounded-prior membership privacy~\cite{yeom2018privacy}, which quantifies the adversary's posterior advantage given a bounded prior on membership and offers a more fine-grained characterization than AUC-ROC alone. We view DP evaluation and refined membership privacy metrics as concrete directions for strengthening the framework's practical applicability.

\section{Conclusions}\label{Conclusion}
As genomic language models (GLMs) are trained and fine-tuned on sensitive cohorts, they risk memorizing individually identifiable sequences, yet no systematic framework has existed to quantify this threat across architectures and training configurations. We addressed this gap with a multi-vector evaluation framework that unifies perplexity-based detection, canary sequence extraction, and membership inference into a maximum vulnerability score characterizing worst-case memorization risk across all attack vectors. Evaluating four GLM architectures across four genomic datasets, we found that all models exhibit measurable memorization under standard fine-tuning, with maximum vulnerability scores revealing sharp architectural differences. Canary extraction success scales monotonically with data duplication, confirming that memorization scaling laws from natural language models transfer to the genomic domain. Crucially, different architectures reveal memorization through different vectors, demonstrating that single-metric evaluations can systematically underestimate privacy exposure. These findings establish an empirical baseline for memorization risk in GLMs and motivate the adoption of multi-vector privacy auditing as a standard practice for genomic AI development.


\bibliographystyle{ACM-Reference-Format}
\bibliography{sample-base}

@inproceedings{zhou2024dnabert2,
  author    = {Zhou, Zhihan and Ji, Yanrong and Li, Weijian and Dutta, Pratik and Davuluri, Ramana V. and Liu, Han},
  title     = {{DNABERT}-2: Efficient Foundation Model and Benchmark for Multi-Species Genome},
  booktitle = {Proceedings of the Twelfth International Conference on Learning Representations (ICLR)},
  year      = {2024}
}

@inproceedings{nguyen2023hyenadna,
  author    = {Nguyen, Eric and Poli, Michael and Faizi, Marjan and Thomas, Armin W. and Wornow, Michael and Birch-Sykes, Callum and Massaroli, Stefano and Patel, Aman and Rabideau, Clayton M. and Bengio, Yoshua and Ermon, Stefano and R\'{e}, Christopher and Baccus, Stephen},
  title     = {{HyenaDNA}: Long-Range Genomic Sequence Modeling at Single Nucleotide Resolution},
  booktitle = {Advances in Neural Information Processing Systems},
  volume    = {36},
  year      = {2023}
}

@article{dallatorre2023nucleotide,
  author    = {Dalla-Torre, Hugo and Gonzalez, Liam and Mendoza-Revilla, Javier and Carranza, Nicolas Lopez and Grzywaczewski, Adam and Ober, Florian and others},
  title     = {The Nucleotide Transformer: Building and Evaluating Robust Foundation Models for Human Genomics},
  journal   = {bioRxiv},
  year      = {2023}
}

@article{nguyen2024evo,
  author    = {Nguyen, Eric and Poli, Michael and Durrant, Matthew G. and Thomas, Armin W. and Kang, Brian and Sullivan, Jeremy and others},
  title     = {Sequence Modeling and Design from Molecular to Genome Scale with {Evo}},
  journal   = {Science},
  volume    = {386},
  number    = {6723},
  pages     = {ado9336},
  year      = {2024}
}

@article{erlich2014routes,
  author    = {Erlich, Yaniv and Narayanan, Arvind},
  title     = {Routes for Breaching and Protecting Genetic Privacy},
  journal   = {Nature Reviews Genetics},
  volume    = {15},
  number    = {6},
  pages     = {409--421},
  year      = {2014}
}

@article{gymrek2013identifying,
  author    = {Gymrek, Melissa and McGuire, Amy L. and Golan, David and Halperin, Eran and Erlich, Yaniv},
  title     = {Identifying Personal Genomes by Surname Inference},
  journal   = {Science},
  volume    = {339},
  number    = {6117},
  pages     = {321--324},
  year      = {2013}
}

@article{homer2008resolving,
  author    = {Homer, Nils and Szelinger, Szabolcs and Redman, Margot and Duggan, David and Tembe, Waibhav and Muehling, Jill and Pearson, John V. and Stephan, Dietrich A. and Nelson, Stanley F. and Craig, David W.},
  title     = {Resolving Individuals Contributing Trace Amounts of {DNA} to Highly Complex Mixtures Using High-Density {SNP} Genotyping Microarrays},
  journal   = {PLoS Genetics},
  volume    = {4},
  number    = {8},
  pages     = {e1000167},
  year      = {2008}
}

@inproceedings{carlini2021extracting,
  author    = {Carlini, Nicholas and Tram\`{e}r, Florian and Wallace, Eric and Jagielski, Matthew and Herbert-Voss, Ariel and Lee, Katherine and Roberts, Adam and Brown, Tom B. and Song, Dawn and Erlingsson, \'{U}lfar and Oprea, Alina and Raffel, Colin},
  title     = {Extracting Training Data from Large Language Models},
  booktitle = {30th USENIX Security Symposium (USENIX Security 21)},
  pages     = {2633--2650},
  year      = {2021}
}

@inproceedings{carlini2019secret,
  author    = {Carlini, Nicholas and Liu, Chang and Erlingsson, \'{U}lfar and Kos, Jernej and Song, Dawn},
  title     = {The Secret Sharer: Evaluating and Testing Unintended Memorization in Neural Networks},
  booktitle = {28th USENIX Security Symposium (USENIX Security 19)},
  pages     = {267--284},
  year      = {2019}
}

@inproceedings{carlini2023quantifying,
  author    = {Carlini, Nicholas and Ippolito, Daphne and Jagielski, Matthew and Lee, Katherine and Tram\`{e}r, Florian and Zhang, Chiyuan},
  title     = {Quantifying Memorization Across Neural Language Models},
  booktitle = {Proceedings of the Eleventh International Conference on Learning Representations (ICLR)},
  year      = {2023}
}

@inproceedings{chen2021dp_genomic,
  author    = {Chen, Junjie and Wang, Xiaoqian and Peng, Bin},
  title     = {Differential Privacy Protection Against Membership Inference Attack on Machine Learning for Genomic Data},
  booktitle = {Pacific Symposium on Biocomputing (PSB)},
  pages     = {26--37},
  year      = {2021}
}

@article{belfiore2025bihmia,
  author    = {Belfiore, Alessandro and Passerat-Palmbach, Jonathan and Usynin, Dmitrii},
  title     = {Biologically-Informed Hybrid Membership Inference Attacks on Generative Genomic Models},
  journal   = {arXiv preprint arXiv:2511.07503},
  year      = {2025}
}

@article{erlich2018identity,
  author    = {Erlich, Yaniv and Shor, Tal and Carmi, Shai and Pe'er, Itsik},
  title     = {Identity Inference of Genomic Data Using Long-Range Familial Searches},
  journal   = {Science},
  volume    = {362},
  number    = {6415},
  pages     = {690--694},
  year      = {2018}
}

@inproceedings{humbert2013kin,
  author    = {Humbert, Mathias and Ayday, Erman and Hubaux, Jean-Pierre and Telenti, Amalio},
  title     = {Addressing the Concerns of the Lacks Family: Quantification of Kin Genomic Privacy},
  booktitle = {Proceedings of the 2013 ACM SIGSAC Conference on Computer \& Communications Security},
  pages     = {1141--1152},
  year      = {2013}
}

@inproceedings{shokri2017membership,
  author    = {Shokri, Reza and Stronati, Marco and Song, Congzheng and Shmatikov, Vitaly},
  title     = {Membership Inference Attacks Against Machine Learning Models},
  booktitle = {Proceedings of the IEEE Symposium on Security and Privacy (S\&P)},
  pages     = {3--18},
  year      = {2017}
}

@inproceedings{yeom2018privacy,
  author    = {Yeom, Samuel and Giacomelli, Irene and Fredrikson, Matt and Jha, Somesh},
  title     = {Privacy Risk in Machine Learning: Analyzing the Connection to Overfitting},
  booktitle = {Proceedings of the IEEE 31st Computer Security Foundations Symposium (CSF)},
  pages     = {268--282},
  year      = {2018}
}

@inproceedings{carlini2022lira,
  author    = {Carlini, Nicholas and Chien, Steve and Nasr, Milad and Song, Shuang and Terzis, Andreas and Tram\`{e}r, Florian},
  title     = {Membership Inference Attacks From First Principles},
  booktitle = {Proceedings of the IEEE Symposium on Security and Privacy (S\&P)},
  pages     = {1897--1914},
  year      = {2022}
}

@inproceedings{duan2024mia_llm,
  author    = {Duan, Michael and Suri, Anshuman and Mireshghallah, Niloofar and Min, Sewon and Shi, Weijia and Zettlemoyer, Luke and Tsvetkov, Yulia and Choi, Yejin and Evans, David and Hajishirzi, Hannaneh},
  title     = {Do Membership Inference Attacks Work on Large Language Models?},
  booktitle = {Conference on Language Modeling (COLM)},
  year      = {2024}
}

@article{hu2022mia_survey,
  author    = {Hu, Hongsheng and Salcic, Zoran and Sun, Lichao and Dobbie, Gillian and Yu, Philip S. and Zhang, Xuyun},
  title     = {Membership Inference Attacks on Machine Learning: A Survey},
  journal   = {ACM Computing Surveys},
  volume    = {54},
  number    = {11s},
  pages     = {1--37},
  year      = {2022}
}

@article{wu2025mia_largescale,
  author    = {Wu, Hao and Cao, Yang},
  title     = {Membership Inference Attacks on Large-Scale Models: A Survey},
  journal   = {arXiv preprint arXiv:2503.19338},
  year      = {2025}
}

@inproceedings{lee2022dedup,
  author    = {Lee, Katherine and Ippolito, Daphne and Nystrom, Andrew and Zhang, Chiyuan and Eck, Douglas and Callison-Burch, Chris and Carlini, Nicholas},
  title     = {Deduplicating Training Data Makes Language Models Better},
  booktitle = {Proceedings of the 60th Annual Meeting of the Association for Computational Linguistics (ACL)},
  pages     = {8424--8445},
  year      = {2022}
}

@inproceedings{kandpal2022dedup_privacy,
  author    = {Kandpal, Nikhil and Wallace, Eric and Raffel, Colin},
  title     = {Deduplicating Training Data Mitigates Privacy Risks in Language Models},
  booktitle = {Proceedings of the 39th International Conference on Machine Learning (ICML)},
  pages     = {10697--10707},
  year      = {2022},
  publisher = {PMLR}
}

@article{almadhoun2020genomic_dp,
  author    = {Almadhoun, Nour and Ayday, Erman and Ulusoy, \"{O}zg\"{u}r},
  title     = {Inference Attacks Against Differentially Private Query Results from Genomic Datasets Including Dependent Tuples},
  journal   = {Bioinformatics},
  volume    = {36},
  number    = {Suppl\_1},
  pages     = {i136--i145},
  year      = {2020}
}

@article{ayoz2021genome_reconstruction,
  author    = {Ayoz, Kerem and Ayday, Erman and Cicek, A. Ercument},
  title     = {Genome Reconstruction Attacks Against Genomic Data-Sharing Beacons},
  journal   = {Proceedings on Privacy Enhancing Technologies (PoPETs)},
  volume    = {2021},
  number    = {3},
  pages     = {28--48},
  year      = {2021}
}

@article{jmir2024genetic_privacy,
  author    = {Faysal, Md Humaion Kabir and Acosta, Juan and Shukla, Harsh and Trivedi, Dhruvil and Khatun, Laila and Kanwal, Shameer},
  title     = {Assessing Privacy Vulnerabilities in Genetic Data Sets},
  journal   = {JMIR Bioinformatics and Biotechnology},
  volume    = {5},
  pages     = {e54332},
  year      = {2024}
}

@inproceedings{li2022dp_learners,
  author    = {Li, Xuechen and Tram\`{e}r, Florian and Liang, Percy and Hashimoto, Tatsunori},
  title     = {Large Language Models Can Be Strong Differentially Private Learners},
  booktitle = {Proceedings of the Tenth International Conference on Learning Representations (ICLR)},
  year      = {2022}
}

@article{kolobkov2024federated_genomic,
  author    = {Kolobkov, Dmitrii and Mishra Sharma, Satyarth and Medvedev, Aleksandr and Lebedev, Mikhail and Kosaretskiy, Egor and Vakhitov, Ruslan},
  title     = {Efficacy of Federated Learning on Genomic Data: A Study on the {UK} Biobank and the 1000 Genomes Project},
  journal   = {Frontiers in Big Data},
  volume    = {7},
  pages     = {1266031},
  year      = {2024}
}

@article{naveed2015privacy_genomic,
  author    = {Naveed, Muhammad and Ayday, Erman and Clayton, Ellen W. and Fellay, Jacques and Gunter, Carl A. and Hubaux, Jean-Pierre and Malin, Bradley A. and Wang, XiaoFeng},
  title     = {Privacy in the Genomic Era},
  journal   = {ACM Computing Surveys},
  volume    = {48},
  number    = {1},
  pages     = {1--44},
  year      = {2015}
}

@article{bonomi2020genomic_sharing,
  author    = {Bonomi, Luca and Huang, Yingxiang and Ohno-Machado, Lucila},
  title     = {Privacy Challenges and Research Opportunities for Genomic Data Sharing},
  journal   = {Nature Genetics},
  volume    = {52},
  number    = {7},
  pages     = {646--654},
  year      = {2020}
}

@inproceedings{fredrikson2015model_inversion,
  author    = {Fredrikson, Matt and Jha, Somesh and Ristenpart, Thomas},
  title     = {Model Inversion Attacks that Exploit Confidence Information and Basic Countermeasures},
  booktitle = {Proceedings of the 22nd ACM SIGSAC Conference on Computer and Communications Security (CCS)},
  pages     = {1322--1333},
  year      = {2015}
}

@inproceedings{mehnaz2022attribute_inference,
  author    = {Mehnaz, Shagufta and Dibbo, Sayanton V. and Kabir, Ehsanul and Li, Ninghui and Bertino, Elisa},
  title     = {Are Your Sensitive Attributes Private? Novel Model Inversion Attribute Inference Attacks on Classification Models},
  booktitle = {31st USENIX Security Symposium (USENIX Security 22)},
  pages     = {4579--4596},
  year      = {2022}
}

@article{hu2022lora,
  title={Lora: Low-rank adaptation of large language models.},
  author={Hu, Edward J and Shen, Yelong and Wallis, Phillip and Allen-Zhu, Zeyuan and Li, Yuanzhi and Wang, Shean and Wang, Liang and Chen, Weizhu and others},
  journal={Iclr},
  volume={1},
  number={2},
  pages={3},
  year={2022}
}

@article{blattner1997complete,
  title={The complete genome sequence of Escherichia coli K-12},
  author={Blattner, Frederick R and Plunkett III, Guy and Bloch, Craig A and Perna, Nicole T and Burland, Valerie and Riley, Monica and Collado-Vides, Julio and Glasner, Jeremy D and Rode, Christopher K and Mayhew, George F and others},
  journal={science},
  volume={277},
  number={5331},
  pages={1453--1462},
  year={1997},
  publisher={American Association for the Advancement of Science}
}

@misc{NCBI_GCF_000005845_2,
  author       = {{NCBI RefSeq}},
  title        = {{Escherichia coli} str. K-12 substr. MG1655, complete genome (ASM584v2, RefSeq assembly GCF\_000005845.2)},
  howpublished = {\url{https://www.ncbi.nlm.nih.gov/datasets/genome/GCF_000005845.2/}},
  note         = {Accessed: 2026-03-04},
  year         = {2026}
}

@article{goffeau1996life,
  title={Life with 6000 genes},
  author={Goffeau, Andr{\'e} and Barrell, Bart G and Bussey, Howard and Davis, Ronald W and Dujon, Bernard and Feldmann, Heinz and Galibert, Francis and Hoheisel, J{\"o}rg D and Jacq, Claude and Johnston, Michael and others},
  journal={Science},
  volume={274},
  number={5287},
  pages={546--567},
  year={1996},
  publisher={American Association for the Advancement of Science}
}

@article{engel2014reference,
  title={The reference genome sequence of Saccharomyces cerevisiae: then and now},
  author={Engel, Stacia R and Dietrich, Fred S and Fisk, Dianna G and Binkley, Gail and Balakrishnan, Rama and Costanzo, Maria C and Dwight, Selina S and Hitz, Benjamin C and Karra, Kalpana and Nash, Robert S and others},
  journal={G3: Genes, Genomes, Genetics},
  volume={4},
  number={3},
  pages={389--398},
  year={2014},
  publisher={Oxford University Press}
}

@misc{NCBI_GCF_000146045_2,
  author       = {{NCBI RefSeq}},
  title        = {{Saccharomyces cerevisiae} S288C reference genome R64 (RefSeq assembly GCF\_000146045.2)},
  howpublished = {\url{https://www.ncbi.nlm.nih.gov/datasets/genome/GCF_000146045.2/}},
  note         = {Accessed: 2026-03-04},
  year         = {2026}
}

@article{nemecek2025exploring,
  title={Exploring Membership Inference Vulnerabilities in Clinical Large Language Models},
  author={Nemecek, Alexander and Yun, Zebin and Rahmani, Zahra and Harel, Yaniv and Chaudhary, Vipin and Sharif, Mahmood and Ayday, Erman},
  journal={arXiv preprint arXiv:2510.18674},
  year={2025}
}

@techreport{nist80030r1,
  author      = {{Joint Task Force Transformation Initiative}},
  title       = {Guide for Conducting Risk Assessments},
  institution = {National Institute of Standards and Technology},
  number      = {NIST SP 800-30 Rev.~1},
  year        = {2012},
  doi         = {10.6028/NIST.SP.800-30r1}
}

@String{Computing = "Computing" }

@String{Computer = "{IEEE} Computer" }

@BOOK{test,
   author = "Donald E. Knuth",
   title = "Seminumerical Algorithms",
   volume = 2,
   series = "The Art of Computer Programming",
   publisher = "Addison-Wesley",
   address = "Reading, MA",
   edition = "2nd",
   month = "10~" # jan,
   year = "1981",
}


\appendix

\section{Training Hyperparameters}
\label{app:hyperparameters}
Table~\ref{tab:hyperparams} reports the full training configurations used for each model. All models use AdamW optimization with linear learning rate warmup and gradient clipping. Early stopping monitors loss on the held-out test split (which also serves as the evaluation target; see Section 4.3 of the main paper) with a minimum improvement threshold of 0.001. For SimpleDNALM, Table~\ref{tab:simplearch} details the architectural hyperparameters of the custom transformer baseline.
\begin{table}[h]
\centering
\small
\caption{SimpleDNALM architectural specification.}
\label{tab:simplearch}
\begin{tabular}{ll}
\toprule
\textbf{Component} & \textbf{Configuration} \\
\midrule
Vocabulary size     & 8 (A, C, G, T + special tokens) \\
Hidden dimension    & 512 \\
Attention heads     & 8 \\
Transformer layers  & 4 \\
Feedforward dimension & 2048 \\
Dropout             & 0.05 \\
Activation          & GELU \\
Positional encoding & Learned embeddings \\
LM head             & Linear (hidden $\rightarrow$ vocab) \\
Loss                & Cross-entropy with label shifting \\
\bottomrule
\end{tabular}
\end{table}

\section{Per-Seed Maximum Vulnerability Analysis}\label{per-seed}
To assess the stability of the maximum vulnerability scores across training runs, we report $S_{\text{config}}$ and $S_{\text{model}}$ disaggregated by seed.

\begin{figure}[h]
\centering
\includegraphics[width=\columnwidth]{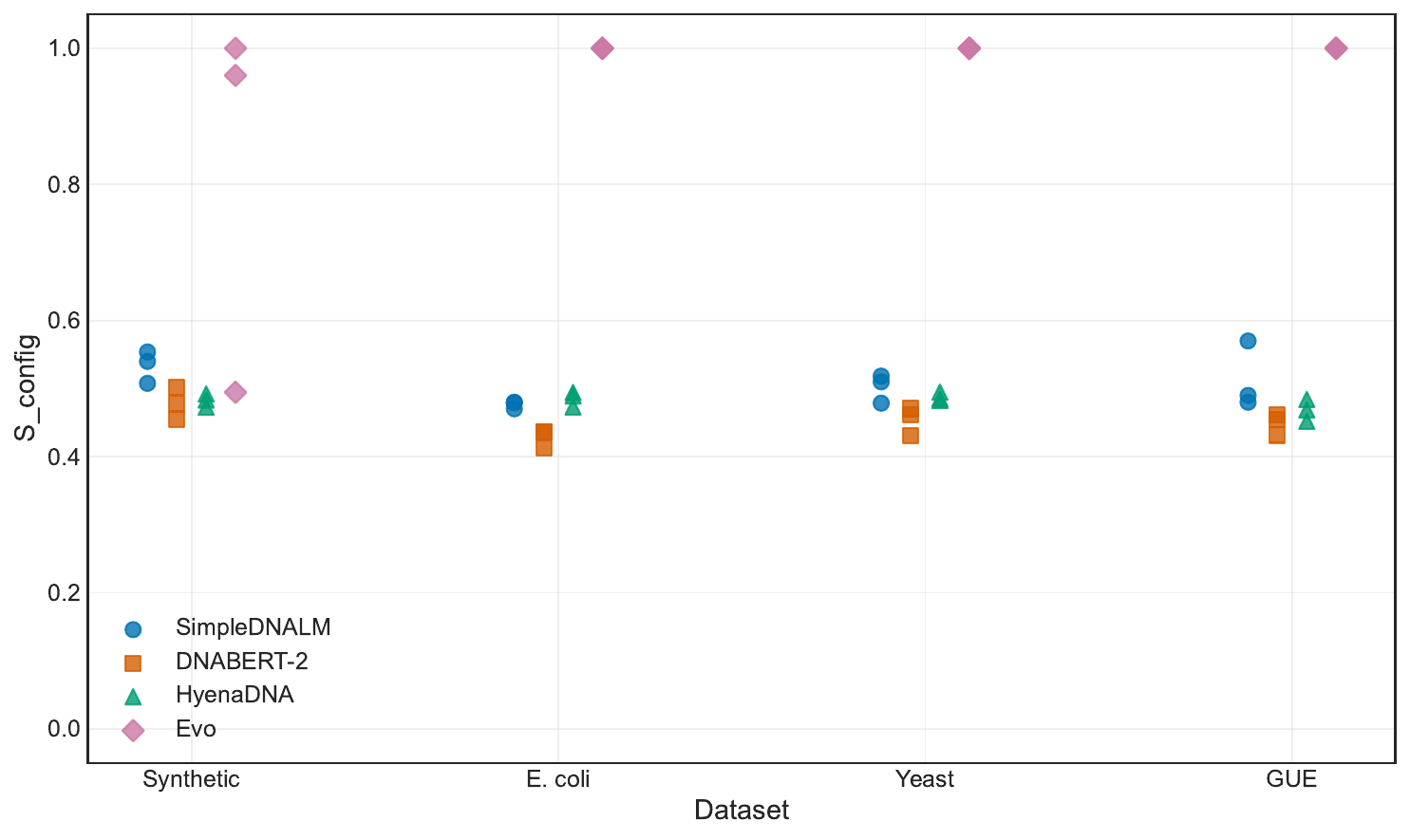}
\caption{$S_{\text{config}}$ per seed for each model-dataset configuration. Each point represents one seed. Evo (LoRA) separates sharply on real genomic datasets while exhibiting high variance on synthetic data. The remaining models cluster tightly with low cross-seed variance.}
\label{fig:sconfig-scatter}
\end{figure}

Figure~\ref{fig:sconfig-scatter} plots $S_{\text{config}}$ for each model-dataset pair across all three seeds. Evo (LoRA) separates clearly from the other architectures on all real genomic datasets, achieving $S_{\text{config}} = 1.00$ across all seeds. On synthetic data, Evo exhibits the only notable cross-seed instability, with one seed producing $S_{\text{config}} = 0.50$ while the other two reach 0.96 and 1.00, reflecting inconsistent extraction behavior in the absence of biological structure. The remaining three models cluster tightly between 0.42 and 0.57 across all configurations, with low cross-seed variance indicating that the vulnerability profiles reported are reproducible.

\begin{table}[h]
\centering
\small
\caption{$S_{\text{model}}$ per seed. Each value represents the maximum $S_{\text{config}}$ across all four datasets for a given model and seed.}
\label{tab:smodel-per-seed}
\begin{tabular}{@{}lcccc@{}}
\toprule
\textbf{Model} & \textbf{Seed 42} & \textbf{Seed 123} & \textbf{Seed 456} & \textbf{Mean $\pm$ Std} \\
\midrule
SimpleDNALM  & 0.55 & 0.57 & 0.54 & 0.55 $\pm$ 0.01 \\
DNABERT-2    & 0.50 & 0.47 & 0.48 & 0.48 $\pm$ 0.01 \\
HyenaDNA     & 0.49 & 0.49 & 0.49 & 0.49 $\pm$ 0.00 \\
Evo (LoRA)   & 1.00 & 1.00 & 1.00 & 1.00 $\pm$ 0.00 \\
\bottomrule
\end{tabular}
\end{table}

Table~\ref{tab:smodel-per-seed} reports $S_{\text{model}}$ for each architecture across all three seeds. Evo achieves the maximum score of 1.00 on every seed, confirming that its extraction vulnerability on real genomic data is not an artifact of a single training run. SimpleDNALM, DNABERT-2, and HyenaDNA all remain within the moderate range with standard deviations of 0.01 or less, indicating stable worst-case behavior across seeds.

To enable independent verification of the aggregation procedure described in Section 5.1 of the main paper, Table~\ref{tab:per-seed-components} reports the full per-seed breakdown of individual component scores ($s_\text{ppl}, s_\text{ext}, s_\text{mia}$) and the resulting $S_\text{config}$ for each model-dataset-seed combination.

\begin{table}[h]
\centering
\footnotesize
\caption{Full results across all models, datasets, and seeds. $S_{\text{config}}$ is the maximum of $s_\text{ppl}$, $s_{\text{ext}}$, and $s_{\text{mia}}$ for each configuration.}
\label{tab:per-seed-components}
\begin{tabular}{@{}llccccc@{}}
\toprule
\textbf{Model} & \textbf{Dataset} & \textbf{Seed} & \textbf{$s_{\text{ppl}}$} & \textbf{$s_{\text{ext}}$} & \textbf{$s_{\text{mia}}$} & \textbf{$S_{\text{config}}$} \\
\midrule
\multirow{12}{*}{SimpleDNALM}
 & \multirow{3}{*}{Synthetic} & 42  & 0.0201  & 0.5300 & 0.5538 & 0.5538 \\
 &                            & 123 & 0.0172  & 0.4300 & 0.5077 & 0.5077 \\
 &                            & 456 & 0.0200  & 0.5400 & 0.5071 & 0.5400 \\
 & \multirow{3}{*}{E. coli}   & 42  & 0.0085  & 0.4500 & 0.4703 & 0.4703 \\
 &                            & 123 & 0.0085  & 0.4100 & 0.4791 & 0.4791 \\
 &                            & 456 & 0.0096  & 0.4800 & 0.4611 & 0.4800 \\
 & \multirow{3}{*}{Yeast}     & 42  & 0.0017  & 0.4800 & 0.5184 & 0.5184 \\
 &                            & 123 & 0.0112  & 0.5100 & 0.4799 & 0.5100 \\
 &                            & 456 & $-$0.0001 & 0.4600 & 0.4786 & 0.4786 \\
 & \multirow{3}{*}{GUE}       & 42  & 0.0045  & 0.4900 & 0.4584 & 0.4900 \\
 &                            & 123 & 0.0076  & 0.5700 & 0.4486 & 0.5700 \\
 &                            & 456 & $-$0.0019 & 0.4800 & 0.4378 & 0.4800 \\
\midrule
\multirow{12}{*}{DNABERT-2}
 & \multirow{3}{*}{Synthetic} & 42  & 0.3400  & 0.2500 & 0.5015 & 0.5015 \\
 &                            & 123 & 0.3477  & 0.1000 & 0.4547 & 0.4547 \\
 &                            & 456 & 0.3631  & 0.0600 & 0.4777 & 0.4777 \\
 & \multirow{3}{*}{E. coli}   & 42  & 0.3448  & 0.1100 & 0.4365 & 0.4365 \\
 &                            & 123 & 0.3447  & 0.1400 & 0.4134 & 0.4134 \\
 &                            & 456 & 0.3287  & 0.1600 & 0.4364 & 0.4364 \\
 & \multirow{3}{*}{Yeast}     & 42  & 0.3715  & 0.1700 & 0.4614 & 0.4614 \\
 &                            & 123 & 0.3776  & 0.1300 & 0.4707 & 0.4707 \\
 &                            & 456 & 0.3899  & 0.2200 & 0.4307 & 0.4307 \\
 & \multirow{3}{*}{GUE}       & 42  & 0.3799  & 0.1100 & 0.4618 & 0.4618 \\
 &                            & 123 & 0.3706  & 0.0700 & 0.4541 & 0.4541 \\
 &                            & 456 & 0.3751  & 0.1000 & 0.4316 & 0.4316 \\
\midrule
\multirow{12}{*}{HyenaDNA}
 & \multirow{3}{*}{Synthetic} & 42  & $-$0.5964 & 0.2700 & 0.4727 & 0.4727 \\
 &                            & 123 & 0.1294    & 0.2500 & 0.4921 & 0.4921 \\
 &                            & 456 & $-$0.6272 & 0.1800 & 0.4833 & 0.4833 \\
 & \multirow{3}{*}{E. coli}   & 42  & $-$0.8547 & 0.1900 & 0.4942 & 0.4942 \\
 &                            & 123 & $-$1.9734 & 0.1100 & 0.4730 & 0.4730 \\
 &                            & 456 & $-$0.2862 & 0.1400 & 0.4890 & 0.4890 \\
 & \multirow{3}{*}{Yeast}     & 42  & $-$0.1289 & 0.1700 & 0.4948 & 0.4948 \\
 &                            & 123 & 0.0353    & 0.1100 & 0.4823 & 0.4823 \\
 &                            & 456 & $-$0.0605 & 0.1600 & 0.4851 & 0.4851 \\
 & \multirow{3}{*}{GUE}       & 42  & $-$0.6826 & 0.2800 & 0.4686 & 0.4686 \\
 &                            & 123 & $-$1.1808 & 0.1700 & 0.4520 & 0.4520 \\
 &                            & 456 & $-$0.8399 & 0.1700 & 0.4841 & 0.4841 \\
\midrule
\multirow{12}{*}{Evo}
 & \multirow{3}{*}{Synthetic} & 42  & 0.0001  & 0.0500 & 0.4946 & 0.4946 \\
 &                            & 123 & 0.2761  & 0.9600 & 0.4448 & 0.9600 \\
 &                            & 456 & 0.3269  & 1.0000 & 0.5018 & 1.0000 \\
 & \multirow{3}{*}{E. coli}   & 42  & 0.3532  & 1.0000 & 0.6271 & 1.0000 \\
 &                            & 123 & 0.3508  & 1.0000 & 0.5673 & 1.0000 \\
 &                            & 456 & 0.3285  & 1.0000 & 0.5254 & 1.0000 \\
 & \multirow{3}{*}{Yeast}     & 42  & 0.3239  & 1.0000 & 0.4884 & 1.0000 \\
 &                            & 123 & 0.3540  & 1.0000 & 0.5074 & 1.0000 \\
 &                            & 456 & 0.3144  & 1.0000 & 0.5019 & 1.0000 \\
 & \multirow{3}{*}{GUE}       & 42  & 0.3392  & 1.0000 & 0.4396 & 1.0000 \\
 &                            & 123 & 0.3381  & 1.0000 & 0.3702 & 1.0000 \\
 &                            & 456 & 0.3355  & 1.0000 & 0.3859 & 1.0000 \\
\bottomrule
\end{tabular}
\end{table}

\section{Canary Detectability Analysis}\label{canary-analysis}
A natural concern with canary-based memorization auditing is whether the inserted canary sequences are themselves detectable by a data holder who inspects the training corpus before model training. If canaries are easily distinguishable from legitimate training sequences, a practitioner could simply filter them out as a preprocessing step, rendering the audit ineffective. We evaluate this scenario by treating canary detection as a binary classification problem over the raw training data.

\textbf{Setup.} We use the same data pipeline and canary insertion procedure as the main experiments. For each of the four datasets (Synthetic, E. coli, Yeast, and GUE) and each of the three random seeds, we load the training sequences and insert 100 canaries at repetition counts of 1, 5, 10, and 20, yielding a mixed corpus of training sequences and canaries. We then compute a set of sequence-level features designed to capture properties that might differ between randomly generated canaries and legitimate training sequences:
\begin{itemize}
    \item Shannon entropy of nucleotide frequencies
    \item GC content (proportion of G$+$C)
    \item Nucleotide frequency variance (uniformity of the ACGT distribution)
    \item Dinucleotide chi-squared statistic (deviation of observed dinucleotide counts from independence, which captures biological constraints such as CpG suppression)
    \item Trinucleotide complexity (fraction of unique 3-mers observed out of 64 possible)
    \item Autocorrelation at lag 1 (Pearson correlation between adjacent nucleotide identities, capturing Markov-like dependencies in biological sequence)
\end{itemize}
\noindent We additionally report sequence length, which is a trivial distinguisher (canaries are 64 nt; training sequences are 256 nt) and is included only for completeness. For each feature, we compute the univariate AUC-ROC as a canary detector. We also train a logistic regression classifier on all features combined using 5-fold cross-validation. All AUC values are reported as mean $\pm$ standard deviation over the three seeds.

\textbf{Results.} Table~\ref{tab:canary-detectability-auc} reports the per-feature and combined classifier AUC for each dataset. Across all four datasets and all features, AUC values fall within a narrow band around 0.50, indicating that no individual feature nor the combined classifier can reliably distinguish canaries from training sequences. This holds not only for the synthetic baseline, where both canaries and training sequences are randomly generated (and indistinguishability is expected), but also for the three real genomic datasets. On E. coli, Yeast, and GUE, features that one might expect to capture biological structure, dinucleotide bias ($\chi^2$), trinucleotide complexity, and autocorrelation all yield AUC $\leq$ 0.51.

\begin{table}[h]
\centering
\small
\caption{Per-feature membership inference AUC scores for the canary detectability analysis across all four evaluation datasets. Each feature is evaluated independently as a membership signal; \texttt{LR\_combined} denotes the likelihood-ratio classifier combining all features. Values reported as mean $\pm$ std across three random seeds.}
\label{tab:canary-detectability-auc}
\begin{tabular}{llc}
\toprule
\textbf{Dataset} & \textbf{Feature} & \textbf{AUC} \\
\midrule
\multirow{8}{*}{Synthetic}
 & \texttt{shannon\_entropy}            & $0.504 \pm 0.001$ \\
 & \texttt{gc\_content}                 & $0.509 \pm 0.012$ \\
 & \texttt{sequence\_length}  & $0.506 \pm 0.005$ \\
 & \texttt{nucleotide\_freq\_variance}  & $0.496 \pm 0.001$ \\
 & \texttt{dinucleotide\_chi2}          & $0.491 \pm 0.015$ \\
 & \texttt{trinucleotide\_complexity}   & $0.491 \pm 0.007$ \\
 & \texttt{autocorrelation\_lag1}       & $0.494 \pm 0.012$ \\
 & \texttt{LR\_combined}               & $0.498 \pm 0.019$ \\
\midrule
\multirow{8}{*}{E. coli}
 & \texttt{shannon\_entropy}            & $0.506 \pm 0.008$ \\
 & \texttt{gc\_content}                 & $0.510 \pm 0.008$ \\
 & \texttt{sequence\_length}  & $0.506 \pm 0.005$ \\
 & \texttt{nucleotide\_freq\_variance}  & $0.494 \pm 0.008$ \\
 & \texttt{dinucleotide\_chi2}          & $0.496 \pm 0.010$ \\
 & \texttt{trinucleotide\_complexity}   & $0.493 \pm 0.008$ \\
 & \texttt{autocorrelation\_lag1}       & $0.492 \pm 0.006$ \\
 & \texttt{LR\_combined}               & $0.504 \pm 0.005$ \\
\midrule
\multirow{8}{*}{Yeast}
 & \texttt{shannon\_entropy}            & $0.510 \pm 0.006$ \\
 & \texttt{gc\_content}                 & $0.511 \pm 0.003$ \\
 & \texttt{sequence\_length}  & $0.506 \pm 0.005$ \\
 & \texttt{nucleotide\_freq\_variance}  & $0.491 \pm 0.005$ \\
 & \texttt{dinucleotide\_chi2}          & $0.485 \pm 0.015$ \\
 & \texttt{trinucleotide\_complexity}   & $0.500 \pm 0.008$ \\
 & \texttt{autocorrelation\_lag1}       & $0.494 \pm 0.014$ \\
 & \texttt{LR\_combined}               & $0.505 \pm 0.019$ \\
\midrule
\multirow{8}{*}{GUE}
 & \texttt{shannon\_entropy}            & $0.502 \pm 0.007$ \\
 & \texttt{gc\_content}                 & $0.501 \pm 0.016$ \\
 & \texttt{sequence\_length}  & $0.506 \pm 0.005$ \\
 & \texttt{nucleotide\_freq\_variance}  & $0.498 \pm 0.007$ \\
 & \texttt{dinucleotide\_chi2}          & $0.500 \pm 0.010$ \\
 & \texttt{trinucleotide\_complexity}   & $0.493 \pm 0.003$ \\
 & \texttt{autocorrelation\_lag1}       & $0.493 \pm 0.015$ \\
 & \texttt{LR\_combined}               & $0.495 \pm 0.027$ \\
\bottomrule
\end{tabular}
\end{table}

\begin{table*}[h]
\centering
\small
\caption{Full training hyperparameters by model. All models use AdamW with weight decay of 0.01, warmup ratio of 0.1, gradient clipping at norm 1.0, and early stopping with patience 5 and $\delta_{\min} = 0.001$.}
\label{tab:hyperparams}
\begin{tabular}{lcccc}
\toprule
\textbf{Hyperparameter} & \textbf{SimpleDNALM} & \textbf{DNABERT-2} & \textbf{HyenaDNA} & \textbf{Evo} \\
\midrule
Max sequence length      & 512          & 512          & 512          & 512          \\
Max epochs               & 50           & 50           & 50           & 25           \\
Batch size               & 8            & 8            & 8            & 4            \\
Learning rate            & $2\mathrm{e}{-5}$ & $2\mathrm{e}{-5}$ & $2\mathrm{e}{-5}$ & $5\mathrm{e}{-5}$ \\
Weight decay             & 0.01         & 0.01         & 0.01         & 0.01         \\
Warmup ratio             & 0.1          & 0.1          & 0.1          & 0.1          \\
Gradient accumulation    & 2            & 2            & 2            & 4            \\
Max gradient norm        & 1.0          & 1.0          & 1.0          & 1.0          \\
Early stopping patience  & 5            & 5            & 5            & 5            \\
\midrule
Fine-tuning strategy     & Full         & Full         & Full         & LoRA         \\
LoRA rank ($r$)          & ---          & ---          & ---          & 8            \\
LoRA scaling ($\alpha$)  & ---          & ---          & ---          & 16           \\
LoRA dropout             & ---          & ---          & ---          & 0.05         \\
LoRA target modules      & ---          & ---          & ---          & \texttt{Wqkv, dense, gated\_layers, wo} \\
\bottomrule
\end{tabular}
\end{table*}

\section{Canary Length Scaling}\label{scaling}
We extend the canary protocol of the main paper (Section 4.3.1) to lengths of 32, 64, and 128 nucleotides, holding all other conditions constant. We exclude canary length 256 because it equals the training sequence length used throughout the paper; at that length, canaries are no longer distinguishable insertions but full training samples, which would change the experimental setup rather than extend the scaling curve. We evaluate SimpleDNALM, DNABERT-2, and HyenaDNA, which together span the architectural paradigms examined in the main text (causal transformer, masked encoder, and long-range convolutional model), on the Synthetic and E.~coli datasets.

\begin{figure}[h]
  \centering
  \includegraphics[width=\linewidth]{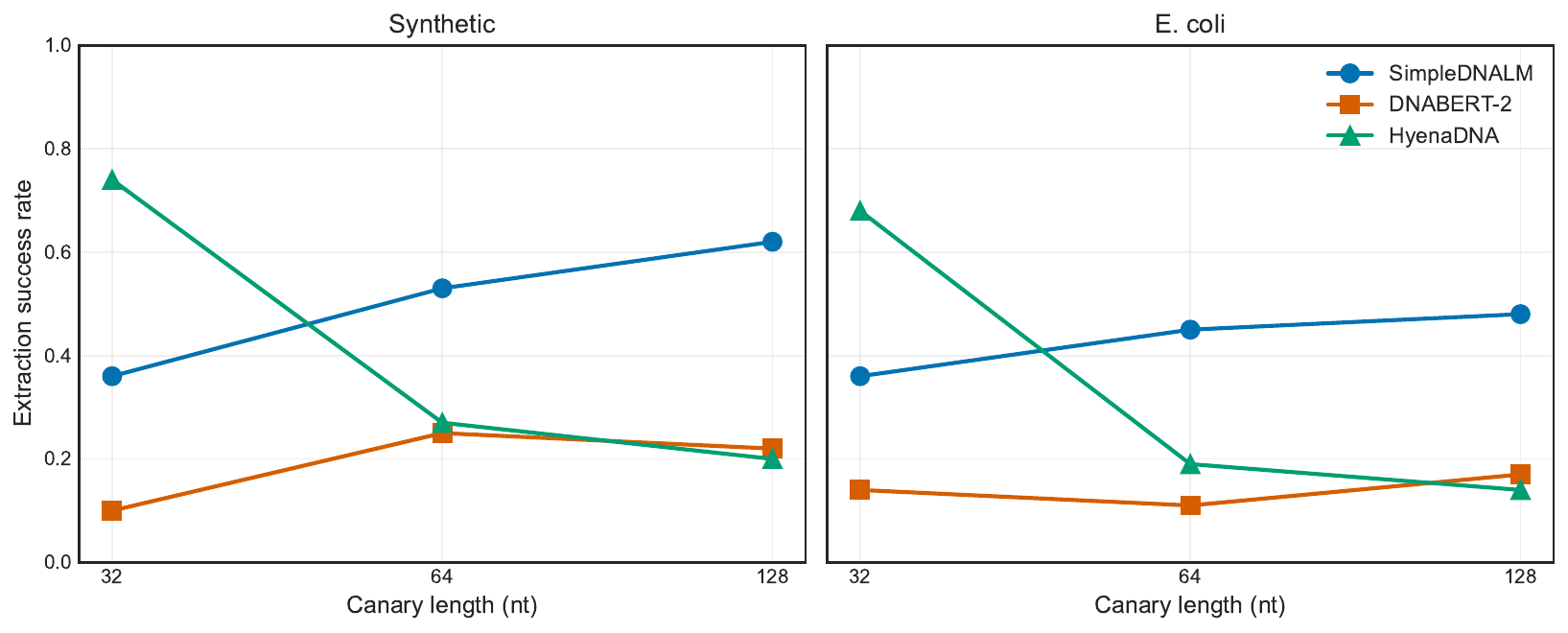}
  \caption{Canary extraction success rate as a function of length.}
  \label{fig:length-scaling}
\end{figure}

Figure~\ref{fig:length-scaling} and Table~\ref{tab:length-scaling} report extraction success rates, perplexity statistics, and membership inference AUC. SimpleDNALM extraction increases monotonically with length (36\%$\rightarrow$62\% on Synthetic; 36\%$\rightarrow$48\% on E.~coli), consistent with longer prefixes providing more conditioning context. DNABERT-2 remains low and flat (10-25\%) while retaining gap ratios above unity (1.49-1.55), preserving the extraction-resistant but loss-detectable profile observed in the main text. HyenaDNA shows the opposite trend, with extraction dropping sharply as length increases (74\%$\rightarrow$20\% on Synthetic; 68\%$\rightarrow$14\% on E.~coli), suggesting that short canaries occupy a small enough search space to be recoverable verbatim while longer ones exceed what the model reproduces reliably.

The qualitative ordering of perplexity gap ratios and membership inference AUC is preserved across lengths, indicating that the architectural memorization profiles characterized in the main text are robust to canary length.

\begin{table}[h]
\centering
\scriptsize
\caption{Perplexity statistics, gap ratio (test/canary), and membership
inference AUC by canary length.}
\label{tab:length-scaling}
\begin{tabular}{llrrrrrr}
\toprule
Model & Dataset & Len & Train & Test & Canary & Gap & MIA \\
\midrule
SimpleDNALM & Synthetic & 32  & 3.92 & 4.00 & 3.92 & 1.02 & 0.75 \\
SimpleDNALM & Synthetic & 64  & 3.93 & 4.00 & 3.92 & 1.02 & 0.78 \\
SimpleDNALM & Synthetic & 128 & 3.96 & 4.01 & 3.96 & 1.01 & 0.76 \\
SimpleDNALM & E.~coli   & 32  & 3.88 & 3.94 & 3.88 & 1.02 & 0.74 \\
SimpleDNALM & E.~coli   & 64  & 3.91 & 3.95 & 3.91 & 1.01 & 0.74 \\
SimpleDNALM & E.~coli   & 128 & 3.94 & 3.96 & 3.94 & 1.00 & 0.75 \\
DNABERT-2   & Synthetic & 32  & 1.79 & 2.75 & 1.80 & 1.53 & 0.75 \\
DNABERT-2   & Synthetic & 64  & 1.75 & 2.65 & 1.75 & 1.52 & 0.75 \\
DNABERT-2   & Synthetic & 128 & 1.81 & 2.68 & 1.80 & 1.49 & 0.75 \\
DNABERT-2   & E.~coli   & 32  & 1.70 & 2.62 & 1.69 & 1.55 & 0.72 \\
DNABERT-2   & E.~coli   & 64  & 1.67 & 2.55 & 1.67 & 1.53 & 0.72 \\
DNABERT-2   & E.~coli   & 128 & 1.65 & 2.50 & 1.65 & 1.52 & 0.72 \\
HyenaDNA    & Synthetic & 32  & 113.09 & 34.84 & 117.16 & 0.30 & 0.73 \\
HyenaDNA    & Synthetic & 64  & 66.55  & 42.21 & 67.38  & 0.63 & 0.74 \\
HyenaDNA    & Synthetic & 128 & 58.84  & 37.48 & 59.61  & 0.63 & 0.74 \\
HyenaDNA    & E.~coli   & 32  & 189.35 & 41.60 & 194.26 & 0.21 & 0.71 \\
HyenaDNA    & E.~coli   & 64  & 26.99  & 14.85 & 27.53  & 0.54 & 0.75 \\
HyenaDNA    & E.~coli   & 128 & 72.09  & 42.57 & 73.33  & 0.58 & 0.75 \\
\bottomrule
\end{tabular}
\end{table}

\end{document}